\documentclass[letterpaper]{article}
\usepackage[preprint]{aaai2027}
\usepackage[hyphens]{url}
\usepackage{graphicx}
\usepackage{natbib}
\usepackage{caption}
\usepackage{amsmath}
\usepackage{amssymb}
\usepackage{amsthm}
\usepackage{booktabs}
\usepackage{multirow}
\usepackage{xcolor}
\usepackage{colortbl}
\usepackage{subfig}
\usepackage{enumitem}
\usepackage[capitalize, noabbrev]{cleveref}
\newcommand{\mathds}[1]{\mathbb{#1}}

\usepackage[most]{tcolorbox}
\newtcolorbox{promptbox}[1][]{
  colback=gray!5!white, colframe=gray!75!black,
  fonttitle=\bfseries, title={System Prompt}, #1}

\crefname{appendix}{Appendix}{Appendices}
\Crefname{appendix}{Appendix}{Appendices}

\title{Label-Free Localization and Test-Time Suppression\\of CLIP's Typographic Circuits}
\author{
    Bohan Liu\textsuperscript{\rm 1}\corresponding,
    Wenqian Ye\textsuperscript{\rm 1},
    Guangzhi Xiong\textsuperscript{\rm 1},
    Zhenghao He\textsuperscript{\rm 1},
    Sanchit Sinha\textsuperscript{\rm 1},
    Aidong Zhang\textsuperscript{\rm 1}\corresponding
}
\affiliations{
    \textsuperscript{\rm 1}University of Virginia\\

    Charlottesville VA 22903, USA\\

    \{qzp4ta, aidong\}@virginia.edu
}

\newcommand{\maincref}[2]{\Cref{#1}}
\begin{document}
\maketitle

\begin{abstract}
CLIP-based Vision-Language Models have become cornerstones of modern multimodal systems. The cross-modal contrastive objective equips the CLIP vision encoder to both see objects and read text. These two abilities induce an entanglement between the encoding of lexical and semantic shapes, creating a distinctive vulnerability to typographic attacks, in which rendered text overrides an image's semantic content with its literal meaning. The black-box nature of these encoders has prevented a mechanistic interpretation of such entanglement. We introduce a head-decomposed Sparse Dictionary Learning (SDL) architecture that, by design, decomposes individual attention heads into interpretable, head-specific latents. Based on this inductive disentanglement, we re-purpose Clarity Score and propose normalized Text Attribution Score, which are both label-free, to identify attention heads that carry typographic information. As a causal validation of successful circuit separation, we perform a mechanistic intervention that reweighs attention away from the patch tokens in the identified lexical heads at test-time. Experiments show that the intervention effectively recovers object classification under typographic attack and, across five CLIP variants and four typographic benchmarks, achieves up to 46.9\% improvement while incurring up to 1\% accuracy loss in general image perception, outperforming prior supervised and training-free methods. Expanding on the mechanistic intervention, we propose a test-time adaptive attention head routing to further reduce the impact on performance at matched robustness, pushing beyond the Pareto frontier of current robustness-perception tradeoff. \looseness=-1
\end{abstract}
\section{Introduction}
\label{sec:intro}

\begin{figure}[t]
\centering
\includegraphics[width=0.8\linewidth]{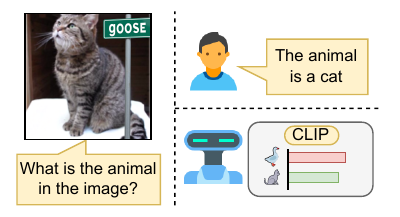}
\caption{\textbf{Motivating example.} A superimposed ``Goose'' text causes CLIP to label a cat as a goose.}
\label{fig:intro}
\end{figure}

Vision encoders trained on web-scale image-text pairs develop two distinct visual processing capacities. 
They encode \textit{semantic shape} for zero-shot recognition, while also encoding \textit{lexical shape} from rendered text, both capacities induced by the contrastive objective in Contrastive Language-Image Pretraining (CLIP)~\citep{radford2021learning}. 
Their entanglement creates a distinctive failure mode: rendered text can distract CLIP models from semantic shape, shifting visual encoding toward the text’s literal meaning: in \Cref{fig:intro}, a cat overlaid with the text ``Goose'' is misclassified as a goose. 

This entanglement, exploited by \textbf{typographic attacks}~\citep{goh2021multimodal}, has been studied in black-box defenses~\citep{denseprefix, disentangle}. The mechanistic question of \textit{where in the network lexical and semantic processing live, and at what computational granularity they are separable,} has remained underexplored. Existing interpretability tools for CLIP either decompose the residual stream into per-layer and per-head components through text-based analysis~\citep{gandelsman2024interpreting, gandelsman2024neurons} or apply Sparse Autoencoders~(SAEs)~\citep{huben2024sparse} to layer outputs as monolithic blocks~\citep{zaigrajew2025interpreting, dreyer2025attributing}. These approaches either fail to leverage the spatial routing of information within an attention head or ignore the division of labor across attention heads, resulting in minimal findings beyond superficial observation of letter-reading attention modules.  Meanwhile, Dyslexify~\citep{dyslexify} investigates the attention weight distribution over text and object shapes, but does not verify what the attended tokens encode in the latent space.

In this paper, we argue that the individual attention head is the key unit for interpreting and resolving such entanglement. Multi-Head Self-Attention~(MHSA) is designed to project the residual stream into parallel low-dimensional subspaces, apply attention separately within each subspace, and recombine via an output projection~\citep{vaswani2017attention}. As descriptive context, head outputs in the last $20\%$ of transformer blocks are approximately mutually decorrelated under Centered Kernel Alignment~\citep{pmlr-v97-kornblith19a} (CKA panels and discussion in the supplement). The primary evidence that the per-head unit is the right granularity is causal and comes from our matched-capacity ablation against a monolithic dictionary (supplement). Therefore, sparse decomposition methods that aggregate across heads contradict this structure, increasing the polysemanticity of learned activations under the sparsity constraints. 

We close this gap with a head-decomposed sparse transcoder that interprets the subspace structure of MHSA. Independent encoders operate on individual head outputs, and a joint decoder reconstructs the post-projection residual stream. The encoder preserves head-local activation by construction, whereas the decoder respects the downstream computation that recombines head latents through the output projection. Our dictionary design provides localized head-level concepts that underpin our circuit mining metrics: \textbf{Clarity Score}, an unsupervised measure of cluster concentration that captures the geometric signature of semantic processing, and \textbf{normalized Text Attribution Score}, which operates on minimally augmented unlabeled images to provide causal concept attribution. They provide a strong signal that separates lexical concepts from semantic concepts without class labels. \looseness=-1

Subsequent circuit mining reveals a separation between lexical and semantic processing among attention heads. Per-head latents partition into two distinct populations: object concepts activate on semantic shapes, and their activating image sets cluster tightly in CLIP space. In contrast, lexical concepts activate on image sets that are visually disparate, held together only by the presence of rendered text. Individual attention heads have concept dictionaries that are overwhelmingly composed of either type. By reweighting attention away from text patches in the identified lexical heads, we causally suppress the lexical circuit while preserving the semantic stream. This mechanistic intervention recovers the semantic representation under typographic attacks, with consistent accuracy gains across five CLIP variants and four typographic benchmarks (up to 46.9\%) and less than 1\% accuracy loss on general clean-data classification. Beyond a practical defense, the result establishes that typographic vulnerability arises from localized entanglement within the CLIP encoder, and disentangling the circuits recovers semantic perception.  

Building up on the mechanistic insights, we propose a test-time adaptive intervention that activates based on individual attention module's output norm, allowing further mitigation of clean-data performance loss and preservation of the model's internal representations. Experiments show that adaptive intervention pushes beyond the Pareto frontier established by prior arts. 
We summarize our contributions as follows:
\begin{itemize}[leftmargin=*,itemsep=0pt,topsep=0pt]
\item \textbf{Head-decomposed sparse transcoders.} We introduce a sparse decomposition architecture that respects the subspace structure of MHSA, motivated by the subspace structure of MHSA, with late-layer head decorrelation as supporting evidence and the mechanistic prerequisite of self-supervised circuit mining signals. 
\item \textbf{Localization of lexical and semantic processing.} We identify the geometric and attributional signals of the two processing streams: we discover that the Clarity Score~\citep{Dreyer2025}, introduced for concept-concentration evaluation, doubles as a head-level robustness indicator, and we propose the normalized Text Attribution Score to separate lexical and semantic attention heads without concept labels.
\item \textbf{Causal intervention recovers semantic perception.} We demonstrate that intervention on identified lexical components recovers object classification under typographic attack across five CLIP ViT variants and four benchmarks, with sub-$1\%$ cost on general classification, outperforming supervised and unsupervised baselines, and an input-conditional adaptive variant that fires only on heads exceeding their clean firing statistics retains the defense at near-zero clean-data accuracy cost on every variant.
\end{itemize}

\section{Related Work}
\label{sec:related}
\subsection{Sparse Dictionary Learning}
Sparse Dictionary Learning (SDL)~\citep{olshausen1997sparse,lee2006efficient} uncovers interpretable structure in high-dimensional representations via sparse linear combinations of basis vectors. Neural architectures such as Sparse Autoencoders~(SAEs)~\citep{huben2024sparse} and Sparse Transcoders~\citep{dunesky2024transcoders} extend SDL by combining reconstruction and sparsity objectives to learn concept dictionaries from model activations in a self-supervised manner, revealing disentangled semantic factors in large vision and language models. However, existing approaches often suffer from polysemanticity due to feature entanglement and a lack of structural constraints~\citep{elhage2022superposition}. Our work introduces a head-decomposed sparse learning architecture that exploits the modularity of MHSA to learn clearer, head-specific semantic dictionaries.
\subsection{Interpretability on Attention Heads}
Multi-Head Self-Attention (MHSA)~\citep{vaswani2017attention} remains underexplored at the level of per-head behavior and semantics. \citet{he2025understandingnatureattentionlowrank} adds unit-channel output and value projections to MHSA to learn sparse concept representations within attention layers. \citet{gandelsman2024interpreting} studies head and layer decomposition in CLIP-ViT by greedily optimizing text descriptions that maximize explained variance per head. Component attribution in this spirit extends beyond CLIP~\citep{balasubramanian2024decomposing}, and \citet{basile2025residual} tie head specialization to the geometry of head representations, likewise reporting late-layer letter-reading heads. \citet{su2025concepts} uses output and value projections into the residual stream~\citep{elhage2021mathematical} to analyze head contributions and concept vectors via SAE~\citep{huben2024sparse}. \citet{kissane2024interpretingattentionlayeroutputs} applies SAEs directly to attention-layer outputs, using decoder-channel norms for attribution, thereby supporting our hypothesis of independent single-head representations. \looseness=-1
\subsection{Disentangling Lexical and Semantic Circuits}
\label{sec:related_compare}
Typographic attacks exploit CLIP's tendency to over-attend to rendered text in images, causing misclassification toward the written label~\citep{goh2021multimodal}. \citet{disentangle} introduces benchmark datasets and demonstrates the separability of visual and lexical processing, but does not localize the responsible units. Defense-Prefix~\citep{denseprefix} learns a text-encoder prefix from annotated datasets to suppress adversarial text. Dyslexify~\citep{dyslexify} greedily searches typographic circuits via attention weights, an approach increasingly noisy as model depth grows. \citet{joseph2025steering} steers CLIP-ViT through SAE features but stops short of identifying the heads causally responsible for lexical processing, and \citet{pach2025sparse} steer multimodal LLMs downstream of CLIP through SAE latents. In contrast, our method is label-free, performs head-decomposed sparse decomposition with causal attribution, and scales across model sizes.

\section{Method}
\begin{figure*}[th]
    \centering
    \includegraphics[width=0.9\linewidth]{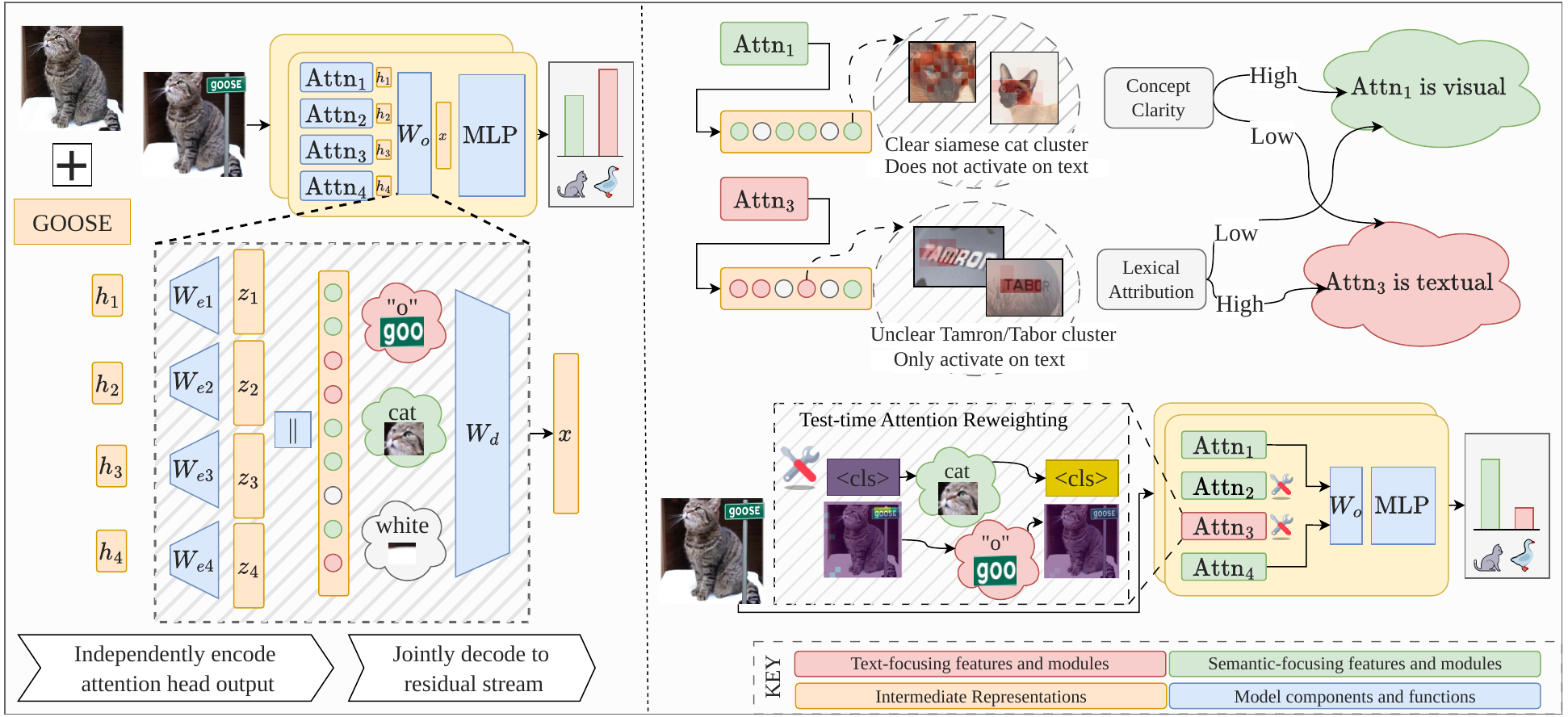}
    \caption{\textbf{Method Overview.} \textbf{(a) Head-decomposed sparse decomposition.} We employ \textbf{Sparse Dictionary Learning} to decompose each attention module ($\text{Attn}_1 \ldots \text{Attn}_4$) into per-head sparse latent neurons. \textbf{(b) Lexical-head localization and causal intervention.} The decomposition surfaces a separation between heads encoding semantic shapes and heads encoding lexical shapes. Identifying the lexical heads (e.g., $\text{Attn}_3$) and reweighting their attention away from patch tokens onto the class token causally suppresses the lexical-processing circuit, recovering semantic perception under typographic attacks.}
    \label{fig:schematic}

\end{figure*}
In this section, we introduce our two-stage method as shown in \Cref{fig:schematic}. \Cref{fig:schematic}(a) presents a head-decomposed sparse transcoder, trained in a self-supervised manner on unlabeled image data, that produces concept-based interpretations of attention modules at per-head granularity. In \Cref{fig:schematic}(b), attention heads are examined based on their corresponding concepts. We leverage two metrics, Clarity Score~(CS) and the proposed Text Attribution Score~(TAS), to disentangle lexical and semantic circuits. Given the transformer circuit composed of the lexical attention heads, the attentions within the circuits are reweighted towards the class token for effective semantic perception under interference. 

\subsection{Preliminaries}

\noindent \textbf{Multi-Head Self-Attention.} Given input sequence $X \in \mathbb{R}^{n \times d_{\text{model}}}$ and projection weights $W_i^Q, W_i^K, W_i^V \in \mathbb{R}^{d_{\text{model}} \times d_{\text{h}}}$ and $W^O \in \mathbb{R}^{(h \cdot d_{\text{h}}) \times d_{\text{model}}}$, an MHSA module with $h$ heads of dimension $d_{\text{h}}$ satisfying $d_{\text{model}} = h \cdot d_{\text{h}}$ computes
\begin{align}
\text{MHSA}(X) &= \text{Concat}(\text{head}_1, \dots, \text{head}_h) W^O, \\
\text{head}_i &= \text{Attn}(XW_i^Q, XW_i^K, XW_i^V),
\label{eq:mhsa}
\end{align}
where $\text{Attn}$ denotes the standard scaled dot-product attention~\citep{vaswani2017attention}.
 
\noindent \textbf{Residual Stream.} Following \citet{gandelsman2024interpreting}, the output $l_i$ of transformer layer $i$ decomposes additively as
\begin{equation}
l_{i} = l_{i-1} + \text{MHSA}(l_{i-1}) + \text{MLP}\!\left(\text{MHSA}(l_{i-1}) + l_{i-1}\right),
\end{equation}
so the final layer output is a sum over per-block attention and MLP contributions. We refer to these contributions and their accumulation $l_i$ as components of the residual stream.

\noindent \textbf{Sparse Autoencoder and Transcoder.} Sparse Autoencoders~(SAEs)~\citep{huben2024sparse} decompose dense representations into a high-dimensional sparse latent space. Given sparse dimension $d_{\text{s}} \gg d_{\text{model}}$, weights $\mathbf{W}_\text{enc} \in \mathbb{R}^{d_{\text{s}}\times d_{\text{model}}}$, $\mathbf{W}_\text{dec} \in \mathbb{R}^{d_{\text{model}}\times d_{\text{s}}}$, biases $\mathbf{b}_{\text{enc}} \in \mathbb{R}^{d_{\text{s}}}, \mathbf{b}_{\text{dec}} \in \mathbb{R}^{d_{\text{model}}}$, and activation $\sigma(\cdot)$, an SAE maps $\mathbf{x} \in \mathbb{R}^{d_\text{model}}$ to
\begin{equation}
    \bar{\mathbf{x}} = \mathbf{W}_\text{dec}\, \sigma(\mathbf{W}_\text{enc} \mathbf{x} + \mathbf{b}_{\text{enc}}) + \mathbf{b}_{\text{dec}},
\end{equation}
trained with reconstruction loss and optional sparsity regularization:
\begin{equation}
\label{eq:sae_loss}
    \mathcal{L}(\mathbf{x}) = \|\mathbf{x} - \bar{\mathbf{x}}\|_2^2 + \lambda \cdot \|\sigma(\mathbf{W}_\text{enc} \mathbf{x} + \mathbf{b}_{\text{enc}})\|_1.
\end{equation}
Sparsity is realized via L1 regularization~\citep{huben2024sparse} or Top-$K$ filtering~\citep{gao2025scaling}, with subsequent variants in both branches~\citep{bussmann2024batchtopk, rajamanoharan2024jumping, zaigrajew2025interpreting, gated-sae}. Transcoders~\citep{dunesky2024transcoders} adapt SAEs by replacing the reconstruction target with a regression target downstream of the input.

\subsection{Interpretability via Attention Head Decomposition}
\label{sec:method_transcoder}
We propose a head-decomposed transcoder that independently encodes each head's output and predicts an attention block's residual stream contribution. Unlike prior work that trains a single encoder on concatenated head outputs~\citep{kissane2024interpretingattentionlayeroutputs}, we train one encoder per head, preventing cross-head entanglement during dictionary learning. Given MHSA output $\mathbf{x} = [\mathbf{x}_1, \ldots, \mathbf{x}_h]$ with each $\mathbf{x}_i \in \mathbb{R}^{d_{\text{h}}}$, per-head encoders $\mathbf{W}_1, \ldots, \mathbf{W}_h \in \mathbb{R}^{d_{\text{hs}}\times d_{\text{h}}}$ with sparse dimension $d_{\text{hs}} \gg d_{\text{h}}$, encoder bias $\mathbf{b}_{\text{enc}} \in \mathbb{R}^{h \cdot d_{\text{hs}}}$, joint decoder $\mathbf{W}_{\text{dec}} \in \mathbb{R}^{(h \cdot d_{\text{h}}) \times (h \cdot d_{\text{hs}})}$, and decoder bias $\mathbf{b}_{\text{dec}} \in \mathbb{R}^{(h \cdot d_{\text{h}})}$, the forward pass is
\begin{align}
    \mathbf{z} = \sigma(\text{Concat}(\mathbf{W}_1 \mathbf{x}_1, \ldots, \mathbf{W}_h \mathbf{x}_h) + \mathbf{b}_{\text{enc}}), \ \bar{\mathbf{x}} = \mathbf{W}_{\text{dec}} \mathbf{z} + \mathbf{b}_{\text{dec}}.
    \label{eq:codec}
\end{align}
The transcoder is trained to predict the post-projection MHSA output $\mathbf{o} = \mathbf{W}_O \mathbf{x} \in \mathbb{R}^{d_{\text{model}}}$ with regression loss and optional L1 sparsity:
\begin{equation}
\label{eq:loss}
    \mathcal{L}(\bar{\mathbf{x}}, \mathbf{o}) = \|\mathbf{o} - \bar{\mathbf{x}}\|_2^2 +  \lambda \cdot \|\mathbf{z}\|_1.
\end{equation}
We formally discuss the structural motivation for this design in the supplemental material.

\subsection{Localizing Lexical and Semantic Circuits}
\label{sec:circuit}
The head-decomposed dictionary shows two geometrically distinct populations of latent concepts. We uncover this separation using two unsupervised metrics: the Clarity Score and the normalized Text Attribution Score, which identify heads in the lexical and semantic circuits without concept labels. We use \emph{lexical heads} and \emph{semantic heads} as shorthand for heads whose latent populations are dominated by lexical and semantic activations, respectively. For a common reading direction, we report the lexicality axis as $-z(\mathrm{CS})$ and $z(\mathrm{nTAS})$, with $z(\cdot)$ denoting per-layer z-standardization, so that higher always denotes stronger lexical enrichment.

\noindent \textbf{Self-supervised Clarity Scoring.} Per-head latents partition into two qualitatively different geometric patterns. Semantic latents activate on visually concentrated image clusters with low mean pairwise distance in CLIP embedding space, as illustrated by the \textit{siamese cat} cluster in \Cref{fig:schematic}. Lexical latents activate on visually disjoint clusters held together only by similar rendered text, as illustrated by the ``\textit{Tamron/Tabor}'' cluster discovered by Attn$_3$ in \Cref{fig:schematic}. We capture this distinction with the Clarity Score~\citep{Dreyer2025}, an unsupervised measure of cluster concentration in CLIP embedding space; proposed originally to evaluate concept concentration in sparse dictionaries, we discover that at the head level it indicates robustness to typographic content and repurpose it for circuit selection. Given a latent $c$ and a set of images $I_c$ that activate $c$, the Clarity Score of $c$ is defined as the average pairwise cosine similarity among the activation set $I_c$ with $|I_c|=N$:
\begin{equation}
    \text{CS}(c) = \frac{1}{N(N-1)} \sum_{i=1}^N \sum_{j=1,j\neq i}^N \frac{\langle\text{CLIP}(I_c^{(i)}), \text{CLIP}(I_c^{(j)})\rangle}{\|\text{CLIP}(I_c^{(i)})\| \cdot \|\text{CLIP}(I_c^{(j)})\|}
\end{equation}
where $\text{CLIP}(\cdot)$ denotes the CLIP image encoder. Because CLIP's contrastive objective aligns embeddings on visual semantic content rather than lexical shape, a higher Clarity Score indicates a more semantically focused concept, and a lower one indicates a lexical concept. Heads whose latents have low Clarity Scores are candidates for the lexical-encoding population. We establish the empirical grounding of this metric in \Cref{sec:converge}.


\noindent \textbf{Causal Text Attribution.}
To complement the geometric signal of CS with a causal one, we quantify the influence of input tokens on each latent activation. Given a ViT input sequence composed of one class token followed by $L$ patch tokens, an attention head of dimension $d_{\text{h}}$, and a target feature defined by the encoder vector $\mathbf{w}_c \in \mathbb{R}^{d_{\text{h}}}$, we first compute the alignment of all $n=L+1$ value vectors $\mathbf{V} \in \mathbb{R}^{n\times d_{\text{h}}}$ with the feature, given by $\mathbf{g}_c = \mathbf{V} \mathbf{w}_c$. Then, the total contribution $b_c$ from the attention head to the feature $c$ is the weighted sum of these alignments over the post-softmax attention weights $\mathbf{a}\in \mathbb{R}^{n}$, given by $b_c =\mathbf{a}^{\top} \mathbf{g}_c$. 
To identify which tokens actively promote this feature, we calculate the gradient of the feature activation $b_c$ with respect to the pre-softmax attention scores $\mathbf{s} \in \mathbb{R}^{n}$. This yields the attribution map
\begin{equation}
\nabla_{\mathbf{s}} b_c = \mathbf{a} \odot (\mathbf{g}_c - \mathbf{a}^{\top} \mathbf{g}_c \cdot \mathbf{1}_n),
\label{eq:attribution_grad}
\end{equation}
where $\odot$ denotes element-wise multiplication and $\mathbf{1}_n = [1, \cdots, 1] \in \mathbb{R}^n$. Intuitively, this gradient assigns positive attribution to tokens whose value contribution $\mathbf{g}_c[j]$ exceeds the attention-weighted average $b_c$, indicating that increased attention to these tokens strengthens the concept's activation. 
Following prior work on attention scoring~\citep{dreyer2025attributing}, we propose a Text Attribution Score that captures how strongly a latent's activation is supported by patches containing rendered text. Given a text-injected image $I_i$ with known text location $M \subseteq \{0, 1, \ldots L\}$, the text-corresponding patch locations yield the text mask  $\mathbf{m}_i \in \mathbb{R}^{L+1}$ with $\mathbf{m}_i[p]=\mathds{1}_M(p)$, where $p$ is the token index and $\mathds{1}_M$ is the indicator function given the set $M$. Then the inverse text mask $\mathbf{m}_i' = \text{Concat}([0],  \mathbf{1}_{n-1} - \mathbf{m}_i)$. The Text Attribution Score~(TAS) and its normalized form~(nTAS) are defined as:
\begin{align}
\label{eq:tas}
    \text{TAS}(c, \mathbf{m}_i) = \frac{\text{ReLU}(\nabla_{\mathbf{s}} b_c) \cdot \mathbf{m}_i}{\langle\mathbf{1}_n, \mathbf{m}_i\rangle}, \\ 
    \text{nTAS}(c, \mathbf{m}_i) = \frac{\text{TAS}(c, \mathbf{m}_i) }{\text{TAS}(c, \mathbf{m}_i)  + \text{TAS}(c, \mathbf{m}_i') }.
\end{align}
In \Cref{eq:tas}, TAS retains only positive gradient attribution on the patch token sequence, matching the implicit ReLU nonlinearity in Top-K transcoders. Normalized TAS jointly considers object and lexical shapes and scales TAS to $[0, 1]$. Per-head scores are computed as the mean over the head's latent concepts. 

In \Cref{sec:exp}, we show that a simple layer-wise statistical threshold on each metric selects the extreme lexical tail for intervention, intermediate heads remain unassigned rather than forced into a binary lexical/semantic ontology, and provides a clean signal for causal intervention to resolve text-shape entanglement, further demonstrating the strength of the mechanistic localization. The augmentation detail for preparing data for nTAS calculation is included in \Cref{sec:augment}.
\subsection{Input-Conditional Head Gating}
\label{sec:gating}
The intervention of \Cref{eq:reweight} acts unconditionally: every selected head is suppressed on every input, including clean images that contain no rendered text. Because the attention head is the bottleneck through which lexical content reaches the class-token representation, the magnitude of a head's output is itself a signal of whether that bottleneck is active on a given input. We therefore propose a simple yet effective input-adaptive intervention. For a selected head $i$ with value vectors $\mathbf{v}_t^{(i)}$, class-token attention weights $\mathbf{a}_{\text{<cls>}}^{(i)}$, and output-projection block $W^O_{(i)} \in \mathbb{R}^{d_{\text{h}} \times d_{\text{model}}}$, define the firing statistic
\begin{equation}
\varphi_i(x) \;=\; \Big\| \Big( \textstyle\sum_{t} \mathbf{a}_{\text{<cls>}}^{(i)}[t]\, \mathbf{v}_t^{(i)}(x) \Big)^{\!\top} W^O_{(i)} \Big\|_2,
\label{eq:firing}
\end{equation}
which is the norm of head $i$'s contribution to the class-token residual on input $x$. Its clean-corpus statistics $(\mu_i, \sigma_i)$ are estimated once on an unlabeled, text-free image corpus. At test time, the head is suppressed only when it fires beyond its clean statistics.
\begin{equation}
g_i(x) \;=\; \mathds{1}\!\left[ \frac{\varphi_i(x) - \mu_i}{\sigma_i} > \theta \right],
\label{eq:gate}
\end{equation}
This adaptive intervention reduces disturbance to ViTs' internal states and only fires when abnormality is detected, keeping clean-data performance intact. The gating statistic at test time is readily available in the forward pass, so applying it induces minimal overhead. 

\section{Experiments}
\label{sec:exp}
\subsection{Implementation Details}
\label{sec:imp_detail}
\noindent \textbf{Evaluation Datasets.}
Our method identifies and disentangles lexical-processing heads from semantic-processing heads within the CLIP encoder; typographic attacks, which directly pit rendered text against visual semantics, are therefore the natural testbed — a successful disentanglement should recover semantic perception precisely where lexical signals would otherwise dominate. The RTA-100~\citep{denseprefix}, Disentangling~\citep{disentangle}, and PAINT~\citep{ilharco2022patching} datasets are used to evaluate the encoder's behavior under typographic attacks, and the Kaggle ImageNet-100 dataset~\citep{shekhar2021imagenet100} is used to evaluate semantic perception on clean data under intervention. Zero-shot object-classification accuracy is the primary metric; zero-shot accuracy on the rendered word (Attack Accuracy) measures the residual lexical influence after intervention. In addition, existing typographic-attack datasets contain either basic text renderings or photos of everyday objects with sticker notes, with limited diversity in objects and text. To obtain a more diverse evaluation, we use Qwen-Image-Edit~\citep{wu2025qwen} to construct a synthetic typographic dataset from Huggingface ImageNet-100~\citep{clane_im100}, denoted ImageNet100-text-D, detailed in \Cref{sec:augment}. \looseness=-1

\noindent \textbf{Baseline Methods.}
We compare our intervention with two prior approaches that also target typographic attacks through different disentanglement approaches: Defense-Prefix~\citep{denseprefix} and Dyslexify~\citep{dyslexify} detailed in \Cref{sec:related_compare}. Both require a class-labeled image dataset for parametric learning or circuit mining, whereas our approach is label-free. Our pipeline also avoids backpropagation through the ViT encoder and replaces per-module forward passes with constant-time filtering over precomputed scores. The computational cost of each approach is summarized in \Cref{app:compute}. All methods are evaluated on ViT-B/16, ViT-L/14, ViT-H/14, ViT-G/14, and ViT-bigG/14, pretrained on LAION-2B~\citep{schuhmann2022laion} and released by OpenCLIP~\citep{cherti2023reproducible}. We additionally report vanilla CLIP under typographic attacks as a no-intervention baseline. 

\noindent \textbf{Sparse Dictionary Learning.} Motivated by the late-layer head-decorrelation analysis and the structural argument (both in the supplement), we attach a Top-128 Head-decomposed Sparse Transcoder with expansion factor 16 to each transformer block in the last $20\%$ of the ViT, where attention heads are approximately decorrelated and individually align with the global semantic embedding. The SDL weights are trained on the ViT model's intermediate representation obtained from the ImageNet training set. The loss function defined in \Cref{eq:loss} is the L2 loss with sparsity coefficient $\lambda=0$ as generally used with Top-K SAEs. Following~\citet{Dreyer2025}, the transcoders are trained for 30 epochs, each epoch using a random-sampled $10\%$ subset of the training set, with the Adam optimizer and cosine learning rate annealing, and an initial learning rate of $1\times 10^{-4}$.

\noindent \textbf{Statistical Head Identification.} Head-level nTAS and CS are obtained by averaging per-latent scores over the ImageNet~\citep{imnet} validation set. Lexical heads are identified by layer-wise statistical thresholding:
\begin{equation}
\label{eq:threshold}
    \text{threshold} = \text{mean}(H_L) + \beta \cdot \text{std}(H_L),
\end{equation}
where $H_L$ is the set of head scores at layer $L$. We use $\beta = -1$ for CS, requiring no validation data, and $\beta > 0$ for nTAS, selected on a synthetic typographic subset based on a 0.1\% sample from ImageNet-100. Heads above the nTAS threshold or below the CS threshold are identified as lexical.

\noindent \textbf{Causal Intervention via Attention Reweighting.} In identified lexical heads, we redirect class-token attention away from patch tokens following Dyslexify~\citep{dyslexify}:
\begin{equation}
\label{eq:reweight}
\mathbf{a}'_{\text{<cls>}}[i] = \begin{cases}
    \alpha & \text{if }i=0  \\
    \mathbf{a}_{\text{<cls>}}[i] \cdot\dfrac{1-\alpha}{\sum\limits_{1\leq j < L+1} \mathbf{a}_{\text{<cls>}}[j]}  & \text{otherwise},
    \end{cases}
\end{equation}
where $\mathbf{a}_{\text{<cls>}} \in \mathbb{R}^{L+1}$ is the original class-token attention map, $\alpha \in [0,1]$ is a control parameter, $i$ indexes tokens ($0$ = \texttt{<cls>}, $1\ldots L$ = patches) following~\citet{dosovitskiy2021an}. We set $\alpha = 1.0$, fully suppressing patch-token contributions in the lexical circuits to demonstrate the accuracy of our circuit mining approach. Full sweep and ablation are in \Cref{app:sweep}.

\subsection{Effect of Intervention on Semantic Perception}
\Cref{tab:compare_regen} evaluates interventions guided by the two metrics of \Cref{sec:circuit}, Clarity Score (CS) and normalized Text Attribution Score (nTAS), against vanilla CLIP under typographic attack. Across all five variants, suppressing the identified lexical heads substantially recovers Object Accuracy on attacked data at sub-point clean cost in most configurations (per-cell confidence intervals in the supplement); the pipeline requires only unlabeled images for SDL training, and CS-based identification is fully label-free. Attack-Accuracy reductions mirror these gains and are tabulated in the supplement.

\begin{table}[th]
    \resizebox{1\linewidth}{!}{
    \begin{tabular}{l|c|c|c|c|c}
\toprule
\textbf{Method} & \textbf{IN-100 (Clean)} & \textbf{RTA-100} & \textbf{Disentangling} & \textbf{PAINT} & \textbf{IN-100-Text-D} \\
\midrule
No intervention & 81.5 & 55.1 & 45.6 & 55.1 & 57.7 \\
Defense-Prefix & \textbf{81.4} (-0.1) & 63.6 & 67.8 & 67.8 & 70.1 \\
Dyslexify & \textit{81.3} (-0.3) & 66.1 & 63.7 & \textit{70.9} & 71.4 \\
Dyslexify$^{*}$ & \color{gray}81.0 (-0.5) & \color{gray}68.5 & \color{gray}70.7 & \color{gray}72.2 & -- \\
Ours~(CS) & 80.5 (-1.0) & \textit{69.9} & \textbf{82.4} & \textit{70.9} & \textit{75.3} \\
Ours~(nTAS) & 81.1 (-0.4) & \textbf{70.8} & \textit{75.8} & \textbf{71.6} & \textbf{76.7} \\
\bottomrule
\end{tabular}

    }
    \centering
    \caption{\textbf{Comparison of intervention strategies, macro-averaged over the five ViT variants}. Numbers are zero-shot Object Accuracy (\%) averaged over ViT-B/16, L/14, H/14, g/14, and bigG/14; ImageNet-100 (Clean) shows the mean difference to the un-intervened model in parentheses. Per-model breakdown in the supplement. Defense-Prefix and Dyslexify rows from their released evaluation artifacts under matched benchmarks. Bold and italic mark first and second place per column. $^{*}$~ indicates numbers reported by the original authors with larger data budget; ``--'' marks benchmarks absent from the original report.}
    \label{tab:compare_regen}
\end{table}
\subsection{Comparison with Prior Works}
\begin{figure}[t]
\centering
\includegraphics[width=0.75\linewidth]{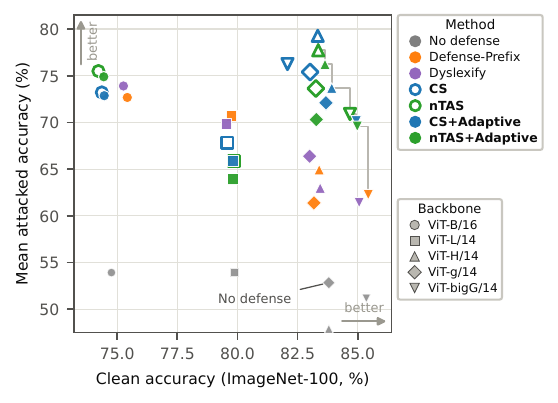}
\caption{\textbf{Robustness--accuracy trade-off.} Mean attacked accuracy (RTA-100, Disentangling, PAINT) vs.\ clean accuracy; gray staircase: Pareto frontier over matched-protocol points. Seven of its eight members are ours.}
\label{fig:pareto}
\end{figure}

We compare against two alternative disentanglement-and-intervention strategies, Defense-Prefix~\citep{denseprefix} and Dyslexify~\citep{dyslexify}. As shown in \Cref{tab:compare_regen}, our intervention achieves the strongest recovery of zero-shot Object Accuracy across typographic-stimulus datasets and model sizes, with only self-supervised SDL and minimal annotated data. Dyslexify~\citep{dyslexify} performs greedy head search by evaluating one (layer, head) module at a time. This approach is constrained by its label requirement and grows noisier as model depth increases. By contrast, our top-down identification requires only constant-time classification evaluations once the lexical circuits are identified, and therefore scales more favorably with model size, exceeding supervised Defense-Prefix~\citep{denseprefix} by a significant margin on the larger ViT backbones. The intervention itself imposes no parameter updates on the ViT encoder: only the head-decomposed transcoder is trained, and its head-wise factorization keeps each per-head encoder proportional to $d_{\text{h}}^2$ rather than the $d_{\text{model}}^2$ of a monolithic encoder. All tested methods show minimal trade-offs in general clean image classification, with macro-averaged accuracy drops of at most $1.1$ points. We further show that adaptive intervention~(\Cref{sec:exp_gating}) reduces cost on clean data accuracy. \Cref{fig:pareto} summarizes the trade-off: seven of the eight Pareto-frontier points over all evaluations belong to our variants, with the adaptive rows extending the frontier's clean-accuracy end. We include more details on the trade-off in the \Cref{app:tradeoff}. Notably, \citet{joseph2025steering} reports perception performance comparable to Defense-Prefix~\citep{denseprefix}, suggesting that SAE-based steering without circuit-level causal analysis plateaus at a similar performance ceiling as black-box prefix tuning. We omit this method from the comparison due to code availability constraints.

\subsection{Input-Conditional Suppression at Zero Clean Cost}
\label{sec:exp_gating}
\begin{table}[th]
    \resizebox{1\linewidth}{!}{
    \begin{tabular}{l|c|c|c|c|c}
\toprule
\textbf{Method} & \textbf{IN-100 (Clean)} & \textbf{RTA-100} & \textbf{Disentangling} & \textbf{PAINT} & \textbf{IN-100-Text-D} \\
\midrule
CS & 80.5 (-1.0) & \textit{69.9} & \textbf{82.4} & \textit{70.9} & 75.3 \\
nTAS & 81.1 (-0.4) & \textbf{70.8} & 75.8 & \textbf{71.6} & \textbf{76.7} \\
CS+Adapt. & \textbf{81.4} (-0.2) & 67.7 & \textit{76.8} & 68.4 & 74.8 \\
nTAS+Adapt. & \textit{81.2} (-0.3) & 69.4 & 73.0 & 70.5 & \textit{76.2} \\
\bottomrule
\end{tabular}

    }
    \centering
    \caption{\textbf{Static versus adaptive intervention, averaged over the five ViT variants.} Bold and italic mark first and second place per column. Gating narrows the clean-data cost while retaining most attacked-accuracy recovery.}
    \label{tab:adaptive}
\end{table}
 The firing statistics $(\mu_i, \sigma_i)$ are calibrated on the clean, unlabeled calibration half of the ImageNet-100 validation set. The threshold is selected once per model by sweeping $\theta \in \{0.25, 0.5, 1.0, 1.5, 2.0, 2.5, 3.0\}$ and maximizing attacked-data accuracy on a held-in half of ImageNet-100-Text-D; all reported results, including the adaptive rows of \Cref{tab:compare_regen}, evaluate the selected $\theta^{*}$ on held-out data under the full-benchmark protocol. Per-model $\theta^{*}$ values are listed in the supplement.
The static intervention suppresses every selected head on every input. The gated variant of \Cref{sec:gating} instead consults each head's firing statistic per image and suppresses only heads that fire beyond their clean-corpus calibration. The CS+Adaptive and nTAS+Adaptive rows of \Cref{tab:compare_regen} evaluate this variant: gating retains the bulk of the static recovery under attack while the clean-data cost falls close to zero across all ten configurations. On ViT-bigG/14, the largest static clean cost of $-2.9$ points falls to $-0.4$ under gating. Selection remains dictionary-scored; only execution is statistics-triggered, so the dictionary is consulted offline, and the runtime overhead is one matrix--vector product and a norm per selected head. 

\subsection{Convergence of Localization Metrics}
\label{sec:converge}
\begin{figure}[t]
\centering
\subfloat[ViT-B/16\label{fig:vitb_corr}]{\includegraphics[width=0.5\linewidth]{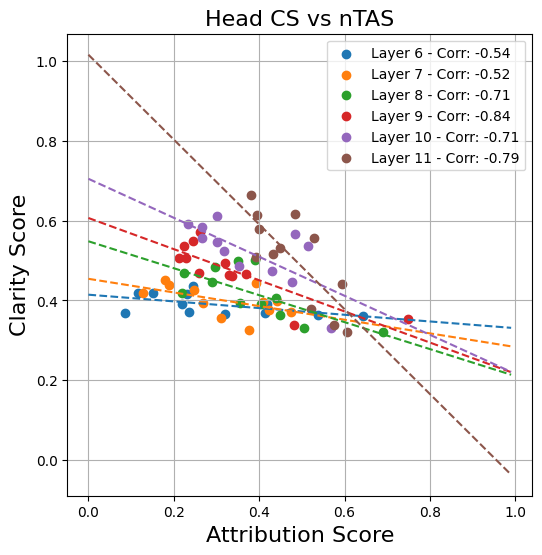}}
\subfloat[ViT-G/14\label{fig:vitg_corr}]{\includegraphics[width=0.5\linewidth]{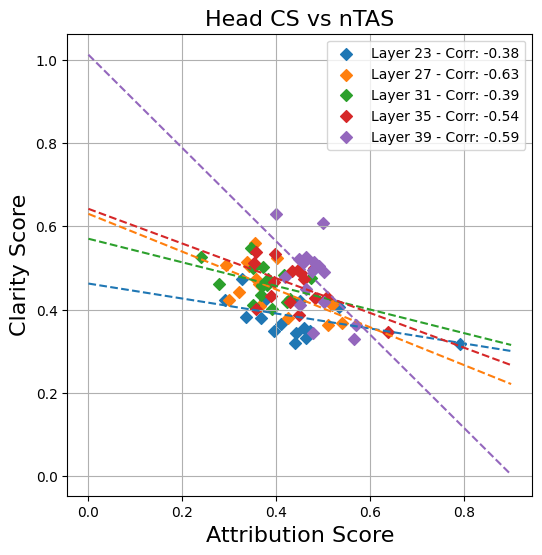}}\hfill \\[-2ex]
\subfloat[Pearson $r$\label{fig:layer_corr}]{\includegraphics[width=0.5\linewidth]{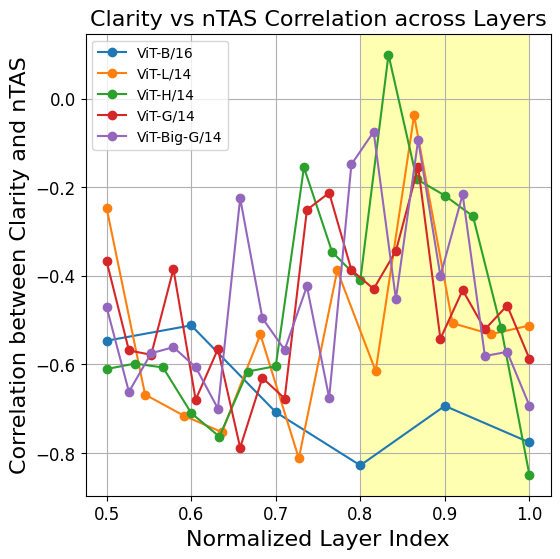}}
\subfloat[Regression slope\label{fig:layer_slope}]{\includegraphics[width=0.5\linewidth]{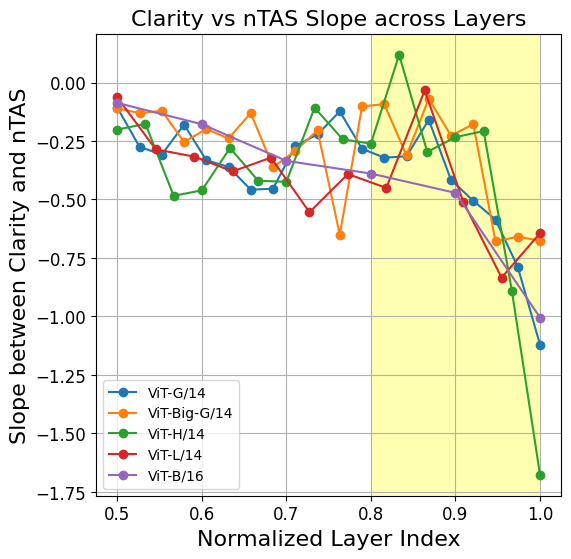}}
\caption{\textbf{CS--nTAS convergence across depth and scale.} \textbf{(a,b)} Per-head CS--nTAS scatter in late layers of ViT-B/16 and ViT-G/14: each point is one attention head; lexical heads cluster at high nTAS / low CS, semantic heads at the opposite corner. \textbf{(c,d)} The anti-correlation strengthens across all CLIP scales: Pearson $r$ becomes more negative, and the regression slope steepens in the final $20\%$ of blocks. }
\label{fig:scatter}\label{fig:corr}
\end{figure}

Plotting per-head $(\text{CS},\, \text{nTAS})$ jointly in the late-layer regime reveals that the two localization metrics, defined independently from concept geometry and from input attribution, identify the same separation. \Cref{fig:scatter} shows that head-level CS and nTAS are strongly anti-correlated in late layers across all tested ViT variants. \Cref{fig:layer_corr,fig:layer_slope} trace this anti-correlation across normalized depth and CLIP scales: the Pearson correlation strengthens sharply, and the regression slope steepens in the same final $20\%$ of blocks. With this correlation, we establish the Clarity Score as a valid signal for disentangling lexical and semantic circuits, given its predictive correlation with the causality-based nTAS metric. \looseness=-1

\subsection{Attribution Faithfulness}
\begin{figure}[t]
    \centering
    \includegraphics[width=0.9\linewidth]{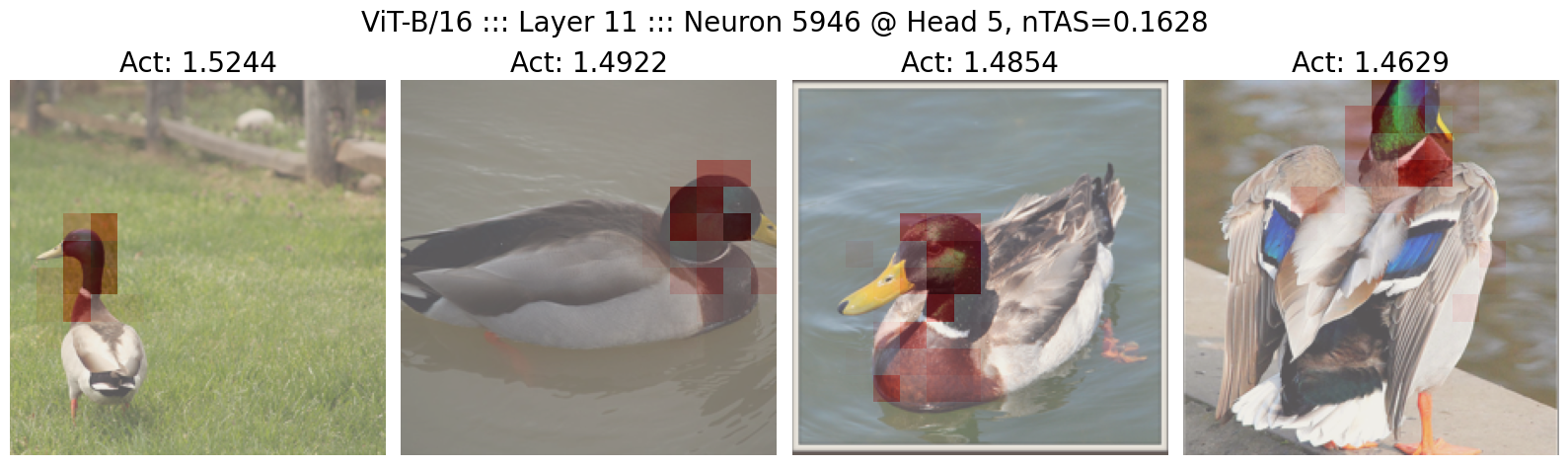}     
    \includegraphics[width=0.9\linewidth]{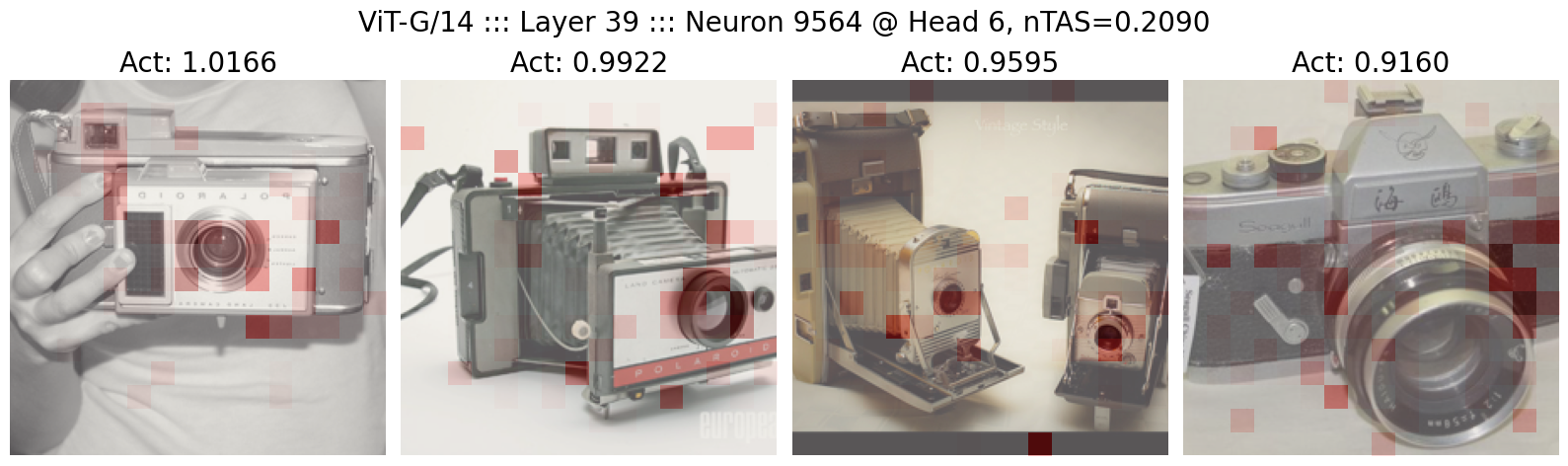}\\[-2ex]
    \noindent\rule{\linewidth}{0.4pt}
    \includegraphics[width=0.9\linewidth]{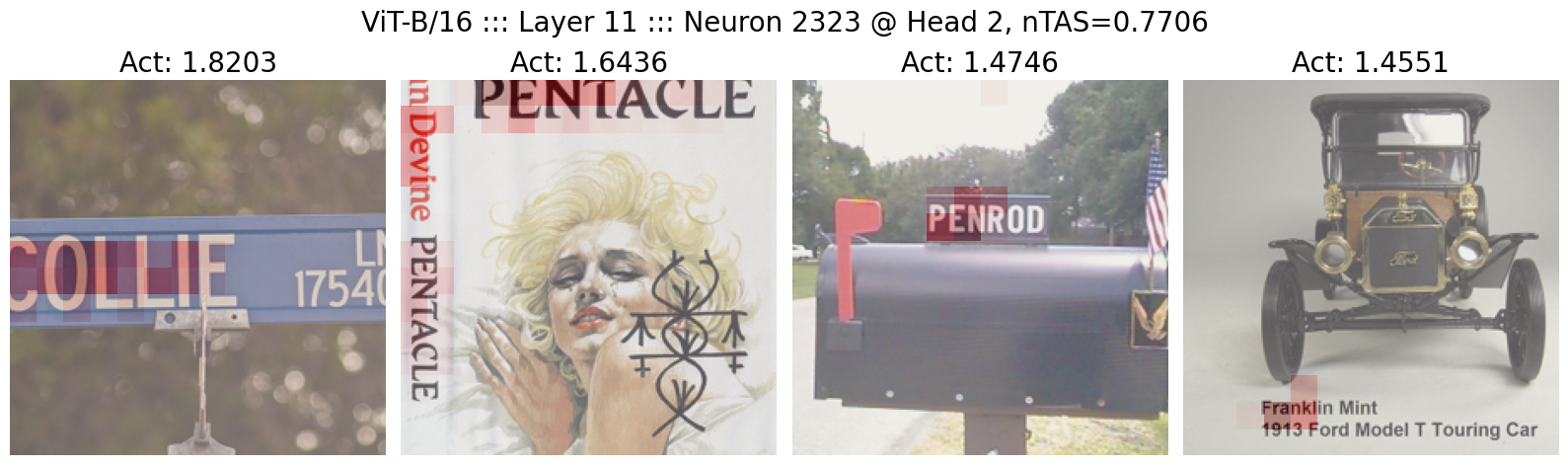}
    \includegraphics[width=0.9\linewidth]{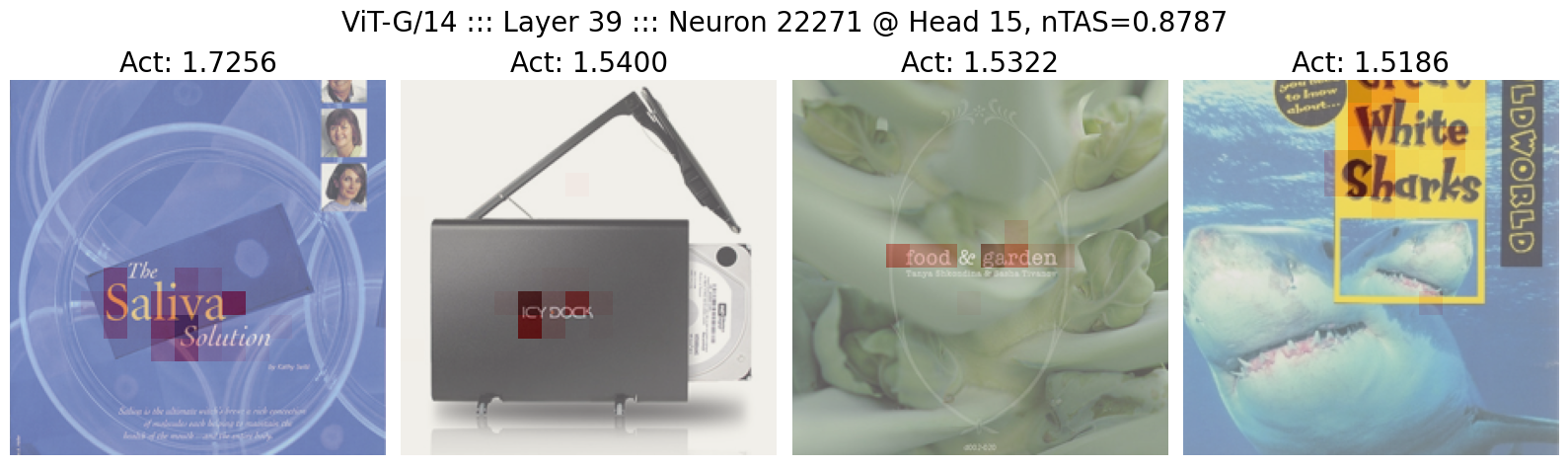}
    \caption{\textbf{Typographic and Semantic Neurons.} Example activating images of neurons and their parent attention head. The top 4 activating images are shown with corresponding neuron attribution overlaid in a red gradient.  Neurons with high nTAS scores are activated by the images that contain text with no significant semantic focus, whereas neurons with low nTAS scores lead to image clusters with strong semantic meaning, e.g., ``Mallard'' in the first row. }
    \label{fig:viz}
\end{figure}

\Cref{fig:viz} provides direct visual evidence for the dual-processing claim. Latents in semantic-encoding heads (low nTAS) activate on visually concentrated image clusters with clear object content (\textit{Mallard} and \textit{Film Camera} in rows 1 and 3), and their gradient attribution maps localize to object regions. Latents in lexical-encoding heads (high nTAS) activate on visually disparate image sets held together only by the presence of rendered text, and their attribution maps localize to the text patches themselves, establishing the causal link between text input and lexical activation. The lexical neurons (ViT-B\#2323, ViT-G\#22271) are tuned to text shapes rather than semantic content, and their activation sets correspondingly receive low Clarity Scores, supporting our proposal that Clarity Score serves as a signal for typographic circuit mining.
More case studies, experiments, and implementation details are available in the Appendix.

\section{Conclusion}
We introduced a head-decomposed sparse transcoder that yields head-attributable concepts by construction, preventing the cross-head entanglement that contaminates monolithic dictionaries. On this decomposition, CS and nTAS identify lexical-encoding heads with minimal labels, and attention reweighting on the identified heads recovers object classification under typographic attack across five CLIP scales and four benchmarks (up to $46.9$ points) at minimal clean cost, surpassing prior supervised and unsupervised defenses. Adaptive intervention conditioned on each head's firing statistics retains this recovery at near-zero clean cost, extending the robustness--perception Pareto frontier. We hope this work pushes the boundary of mechanistic interpretability in multimodal perception systems. \looseness=-1

\bibliography{neurips_original,additions}

@inproceedings{
balasubramanian2024decomposing,
title={Decomposing and Interpreting Image Representations via Text in ViTs Beyond {CLIP}},
author={Sriram Balasubramanian and Samyadeep Basu and Soheil Feizi},
booktitle={The Thirty-eighth Annual Conference on Neural Information Processing Systems},
year={2024},
url={https://openreview.net/forum?id=Vhh7ONtfvV}
}

@article{
basile2025residual,
title={ResiDual Transformer Alignment with Spectral Decomposition},
author={Lorenzo Basile and Valentino Maiorca and Luca Bortolussi and Emanuele Rodol{\`a} and Francesco Locatello},
journal={Transactions on Machine Learning Research},
issn={2835-8856},
year={2025},
url={https://openreview.net/forum?id=z37LCgSIzI},
note={}
}

@inproceedings{cherti2023reproducible,
  title={Reproducible scaling laws for contrastive language-image learning},
  author={Cherti, Mehdi and Beaumont, Romain and Wightman, Ross and Wortsman, Mitchell and Ilharco, Gabriel and Gordon, Cade and Schuhmann, Christoph and Schmidt, Ludwig and Jitsev, Jenia},
  booktitle={Proceedings of the IEEE/CVF conference on computer vision and pattern recognition},
  pages={2818--2829},
  year={2023}
}

@inproceedings{pach2025sparse,
title={Sparse Autoencoders Learn Monosemantic Features in Vision-Language Models},
author={Mateusz Pach and Shyamgopal Karthik and Quentin Bouniot and Serge Belongie and Zeynep Akata},
booktitle={The Thirty-ninth Annual Conference on Neural Information Processing Systems},
year={2025},
url={https://openreview.net/forum?id=DaNnkQJSQf}
}

@String(CVPR= {IEEE Conf. Comput. Vis. Pattern Recog.})

@String(ECCV= {Eur. Conf. Comput. Vis.})

@String(CVPR  = {CVPR})

@String(ECCV  = {ECCV})

@article{gemmateam2025gemma3technicalreport,
      title={Gemma 3 Technical Report}, 
      author={Gemma Team and Aishwarya Kamath and Johan Ferret and Shreya Pathak and Nino Vieillard and Ramona Merhej and Sarah Perrin and Tatiana Matejovicova and Alexandre Ramé and Morgane Rivière and Louis Rouillard and Thomas Mesnard and Geoffrey Cideron and Jean-bastien Grill and Sabela Ramos and Edouard Yvinec and Michelle Casbon and Etienne Pot and Ivo Penchev and Gaël Liu and Francesco Visin and Kathleen Kenealy and Lucas Beyer and Xiaohai Zhai and Anton Tsitsulin and Robert Busa-Fekete and Alex Feng and Noveen Sachdeva and Benjamin Coleman and Yi Gao and Basil Mustafa and Iain Barr and Emilio Parisotto and David Tian and Matan Eyal and Colin Cherry and Jan-Thorsten Peter and Danila Sinopalnikov and Surya Bhupatiraju and Rishabh Agarwal and Mehran Kazemi and Dan Malkin and Ravin Kumar and David Vilar and Idan Brusilovsky and Jiaming Luo and Andreas Steiner and Abe Friesen and Abhanshu Sharma and Abheesht Sharma and Adi Mayrav Gilady and Adrian Goedeckemeyer and Alaa Saade and Alex Feng and Alexander Kolesnikov and Alexei Bendebury and Alvin Abdagic and Amit Vadi and András György and André Susano Pinto and Anil Das and Ankur Bapna and Antoine Miech and Antoine Yang and Antonia Paterson and Ashish Shenoy and Ayan Chakrabarti and Bilal Piot and Bo Wu and Bobak Shahriari and Bryce Petrini and Charlie Chen and Charline Le Lan and Christopher A. Choquette-Choo and CJ Carey and Cormac Brick and Daniel Deutsch and Danielle Eisenbud and Dee Cattle and Derek Cheng and Dimitris Paparas and Divyashree Shivakumar Sreepathihalli and Doug Reid and Dustin Tran and Dustin Zelle and Eric Noland and Erwin Huizenga and Eugene Kharitonov and Frederick Liu and Gagik Amirkhanyan and Glenn Cameron and Hadi Hashemi and Hanna Klimczak-Plucińska and Harman Singh and Harsh Mehta and Harshal Tushar Lehri and Hussein Hazimeh and Ian Ballantyne and Idan Szpektor and Ivan Nardini and Jean Pouget-Abadie and Jetha Chan and Joe Stanton and John Wieting and Jonathan Lai and Jordi Orbay and Joseph Fernandez and Josh Newlan and Ju-yeong Ji and Jyotinder Singh and Kat Black and Kathy Yu and Kevin Hui and Kiran Vodrahalli and Klaus Greff and Linhai Qiu and Marcella Valentine and Marina Coelho and Marvin Ritter and Matt Hoffman and Matthew Watson and Mayank Chaturvedi and Michael Moynihan and Min Ma and Nabila Babar and Natasha Noy and Nathan Byrd and Nick Roy and Nikola Momchev and Nilay Chauhan and Noveen Sachdeva and Oskar Bunyan and Pankil Botarda and Paul Caron and Paul Kishan Rubenstein and Phil Culliton and Philipp Schmid and Pier Giuseppe Sessa and Pingmei Xu and Piotr Stanczyk and Pouya Tafti and Rakesh Shivanna and Renjie Wu and Renke Pan and Reza Rokni and Rob Willoughby and Rohith Vallu and Ryan Mullins and Sammy Jerome and Sara Smoot and Sertan Girgin and Shariq Iqbal and Shashir Reddy and Shruti Sheth and Siim Põder and Sijal Bhatnagar and Sindhu Raghuram Panyam and Sivan Eiger and Susan Zhang and Tianqi Liu and Trevor Yacovone and Tyler Liechty and Uday Kalra and Utku Evci and Vedant Misra and Vincent Roseberry and Vlad Feinberg and Vlad Kolesnikov and Woohyun Han and Woosuk Kwon and Xi Chen and Yinlam Chow and Yuvein Zhu and Zichuan Wei and Zoltan Egyed and Victor Cotruta and Minh Giang and Phoebe Kirk and Anand Rao and Kat Black and Nabila Babar and Jessica Lo and Erica Moreira and Luiz Gustavo Martins and Omar Sanseviero and Lucas Gonzalez and Zach Gleicher and Tris Warkentin and Vahab Mirrokni and Evan Senter and Eli Collins and Joelle Barral and Zoubin Ghahramani and Raia Hadsell and Yossi Matias and D. Sculley and Slav Petrov and Noah Fiedel and Noam Shazeer and Oriol Vinyals and Jeff Dean and Demis Hassabis and Koray Kavukcuoglu and Clement Farabet and Elena Buchatskaya and Jean-Baptiste Alayrac and Rohan Anil and Dmitry and Lepikhin and Sebastian Borgeaud and Olivier Bachem and Armand Joulin and Alek Andreev and Cassidy Hardin and Robert Dadashi and Léonard Hussenot},
      year={2025},
      journal={arXiv preprint arXiv:2503.19786},
      eprint={2503.19786},
      archivePrefix={arXiv},
      primaryClass={cs.CL},
      url={https://arxiv.org/abs/2503.19786}, 
}

@article{waseda2025read,
    title        = {Read or Ignore? A Unified Benchmark for Typographic-Attack Robustness and Text Recognition in Vision-Language Models},
    author       = {Waseda, Futa and Yamabe, Shojiro and Shiono, Daiki and Sasaki, Kento and Takahashi, Tsubasa},
    year         = {2025},
    eprint       = {2512.11899},
    archivePrefix= {arXiv},
    journal={arXiv preprint arXiv:2512.11899},
    primaryClass = {cs.CV},
    url          = {https://arxiv.org/abs/2512.11899},
}

@inproceedings{joseph2025steering,
  title={Steering CLIP's vision transformer with sparse autoencoders},
  author={Joseph, Sonia and Suresh, Praneet and Goldfarb, Ethan and Hufe, Lorenz and Gandelsman, Yossi and Graham, Robert and Bzdok, Danilo and Samek, Wojciech and Richards, Blake Aaron},
  booktitle={Mechanistic Interpretability for Vision at CVPR 2025 (Non-proceedings Track)},
  year={2025}
}

@inproceedings{
dosovitskiy2021an,
title={An Image is Worth 16x16 Words: Transformers for Image Recognition at Scale},
author={Alexey Dosovitskiy and Lucas Beyer and Alexander Kolesnikov and Dirk Weissenborn and Xiaohua Zhai and Thomas Unterthiner and Mostafa Dehghani and Matthias Minderer and Georg Heigold and Sylvain Gelly and Jakob Uszkoreit and Neil Houlsby},
booktitle={International Conference on Learning Representations},
year={2021},
url={https://openreview.net/forum?id=YicbFdNTTy}
}

@article{wang2025internvl3,
  title={Internvl3. 5: Advancing open-source multimodal models in versatility, reasoning, and efficiency},
  author={Wang, Weiyun and Gao, Zhangwei and Gu, Lixin and Pu, Hengjun and Cui, Long and Wei, Xingguang and Liu, Zhaoyang and Jing, Linglin and Ye, Shenglong and Shao, Jie and others},
  journal={arXiv preprint arXiv:2508.18265},
  year={2025}
}

@article{bai2025qwen3,
  title={Qwen3-vl technical report},
  author={Bai, Shuai and Cai, Yuxuan and Chen, Ruizhe and Chen, Keqin and Chen, Xionghui and Cheng, Zesen and Deng, Lianghao and Ding, Wei and Gao, Chang and Ge, Chunjiang and others},
  journal={arXiv preprint arXiv:2511.21631},
  year={2025}
}

@inproceedings{
bussmann2024batchtopk,
title={BatchTopK Sparse Autoencoders},
author={Bart Bussmann and Patrick Leask and Neel Nanda},
booktitle={NeurIPS 2024 Workshop on Scientific Methods for Understanding Deep Learning},
year={2024},
url={https://openreview.net/forum?id=d4dpOCqybL}
}

@INPROCEEDINGS{imnet,
  author={Deng, Jia and Dong, Wei and Socher, Richard and Li, Li-Jia and Kai Li and Li Fei-Fei},
  booktitle={2009 IEEE Conference on Computer Vision and Pattern Recognition}, 
  title={ImageNet: A large-scale hierarchical image database}, 
  year={2009},
  volume={},
  number={},
  pages={248-255},
  doi={10.1109/CVPR.2009.5206848}}

@InProceedings{pmlr-v97-kornblith19a,
  title = 	 {Similarity of Neural Network Representations Revisited},
  author =       {Kornblith, Simon and Norouzi, Mohammad and Lee, Honglak and Hinton, Geoffrey},
  booktitle = 	 {Proceedings of the 36th International Conference on Machine Learning},
  pages = 	 {3519--3529},
  year = 	 {2019},
  editor = 	 {Chaudhuri, Kamalika and Salakhutdinov, Ruslan},
  volume = 	 {97},
  series = 	 {Proceedings of Machine Learning Research},
  month = 	 {09--15 Jun},
  publisher =    {PMLR},
  url = 	 {https://proceedings.mlr.press/v97/kornblith19a.html}
}

@article{rajamanoharan2024jumping,
  title={Jumping ahead: Improving reconstruction fidelity with jumprelu sparse autoencoders},
  author={Rajamanoharan, Senthooran and Lieberum, Tom and Sonnerat, Nicolas and Conmy, Arthur and Varma, Vikrant and Kram{\'a}r, J{\'a}nos and Nanda, Neel},
  journal={arXiv preprint arXiv:2407.14435},
  year={2024}
}

@inproceedings{
zaigrajew2025interpreting,
title={Interpreting {CLIP} with Hierarchical Sparse Autoencoders},
author={Vladimir Zaigrajew and Hubert Baniecki and Przemyslaw Biecek},
booktitle={Forty-second International Conference on Machine Learning},
year={2025},
url={https://openreview.net/forum?id=5MQQsenQBm}
}

@article{gandelsman2024neurons,
      title={Interpreting the Second-Order Effects of Neurons in CLIP}, 
      author={Yossi Gandelsman and Alexei A. Efros and Jacob Steinhardt},
      year={2024},
      journal={arXiv preprint arXiv:2406.04341},
      primaryClass={cs.CV},
      url={https://arxiv.org/abs/2406.04341}, 
}

@misc{shekhar2021imagenet100,
  author       = {Shekhar, Ambesh},
  title        = {ImageNet100},
  year         = {2021},
  howpublished = {Kaggle},
  url          = {https://www.kaggle.com/datasets/ambityga/imagenet100},
  note         = {Accessed: 2026-01-16}
}

@inproceedings{clane_im100,
author = {Tian, Yonglong and Krishnan, Dilip and Isola, Phillip},
title = {Contrastive Multiview Coding},
year = {2020},
isbn = {978-3-030-58620-1},
publisher = {Springer-Verlag},
address = {Berlin, Heidelberg},
url = {https://doi.org/10.1007/978-3-030-58621-8_45},
doi = {10.1007/978-3-030-58621-8_45},
booktitle = {Computer Vision – ECCV 2020: 16th European Conference, Glasgow, UK, August 23–28, 2020, Proceedings, Part XI},
pages = {776–794},
numpages = {19},
location = {Glasgow, United Kingdom}
}

@InProceedings{disentangle,
    author    = {Materzy\'nska, Joanna and Torralba, Antonio and Bau, David},
    title     = {Disentangling Visual and Written Concepts in CLIP},
    booktitle = {Proceedings of the IEEE/CVF Conference on Computer Vision and Pattern Recognition (CVPR)},
    month     = {June},
    year      = {2022},
    pages     = {16410-16419}
}

@article{goh2021multimodal,
  title={Multimodal Neurons in Artificial Neural Networks},
  author={Goh, Gabriel and Cammarata, Nick and Voss, Chelsea and Carter, Shan and Petrov, Michael and Schubert, Ludwig and Radford, Alec and Olah, Chris},
  journal={Distill},
  volume={6},
  number={3},
  pages={e30},
  year={2021}
}

@Article{Dreyer2025,
author={Dreyer, Maximilian
and Berend, Jim
and Labarta, Tobias
and Vielhaben, Johanna
and Wiegand, Thomas
and Lapuschkin, Sebastian
and Samek, Wojciech},
title={Mechanistic understanding and validation of large AI models with SemanticLens},
journal={Nature Machine Intelligence},
year={2025},
month={Sep},
day={01},
volume={7},
number={9},
pages={1572-1585},
issn={2522-5839},
doi={10.1038/s42256-025-01084-w},
url={https://doi.org/10.1038/s42256-025-01084-w}
}

@inproceedings{
gao2025scaling,
title={Scaling and evaluating sparse autoencoders},
author={Leo Gao and Tom Dupre la Tour and Henk Tillman and Gabriel Goh and Rajan Troll and Alec Radford and Ilya Sutskever and Jan Leike and Jeffrey Wu},
booktitle={The Thirteenth International Conference on Learning Representations},
year={2025},
url={https://openreview.net/forum?id=tcsZt9ZNKD}
}

@article{dreyer2025attributing,
  title={From What to How: Attributing CLIP's Latent Components Reveals Unexpected Semantic Reliance},
  author={Dreyer, Maximilian and Hufe, Lorenz and Berend, Jim and Wiegand, Thomas and Lapuschkin, Sebastian and Samek, Wojciech},
  journal={arXiv preprint arXiv:2505.20229},
  year={2025}
}

@InProceedings{radford2021learning,
  title = 	 {Learning Transferable Visual Models From Natural Language Supervision},
  author =       {Radford, Alec and Kim, Jong Wook and Hallacy, Chris and Ramesh, Aditya and Goh, Gabriel and Agarwal, Sandhini and Sastry, Girish and Askell, Amanda and Mishkin, Pamela and Clark, Jack and Krueger, Gretchen and Sutskever, Ilya},
  booktitle = 	 {Proceedings of the 38th International Conference on Machine Learning},
  pages = 	 {8748--8763},
  year = 	 {2021},
  editor = 	 {Meila, Marina and Zhang, Tong},
  volume = 	 {139},
  series = 	 {Proceedings of Machine Learning Research},
  month = 	 {18--24 Jul},
  publisher =    {PMLR},
  url = 	 {https://proceedings.mlr.press/v139/radford21a.html}
}

@article{kissane2024interpretingattentionlayeroutputs,
      title={Interpreting Attention Layer Outputs with Sparse Autoencoders}, 
      author={Connor Kissane and Robert Krzyzanowski and Joseph Isaac Bloom and Arthur Conmy and Neel Nanda},
      year={2024},
      eprint={2406.17759},
      journal={arXiv preprint arXiv:2406.17759},
      archivePrefix={arXiv},
      primaryClass={cs.LG},
      url={https://arxiv.org/abs/2406.17759}, 
}

@inproceedings{vaswani2017attention,
 author = {Vaswani, Ashish and Shazeer, Noam and Parmar, Niki and Uszkoreit, Jakob and Jones, Llion and Gomez, Aidan N and Kaiser, \L ukasz and Polosukhin, Illia},
 booktitle = {Advances in Neural Information Processing Systems},
 editor = {I. Guyon and U. Von Luxburg and S. Bengio and H. Wallach and R. Fergus and S. Vishwanathan and R. Garnett},
 pages = {},
 publisher = {Curran Associates, Inc.},
 title = {Attention is All you Need},
 url = {https://proceedings.neurips.cc/paper_files/paper/2017/file/3f5ee243547dee91fbd053c1c4a845aa-Paper.pdf},
 volume = {30},
 year = {2017}
}

@article{elhage2021mathematical,
   title={A Mathematical Framework for Transformer Circuits},
   author={Elhage, Nelson and Nanda, Neel and Olsson, Catherine and Henighan, Tom and Joseph, Nicholas and Mann, Ben and Askell, Amanda and Bai, Yuntao and Chen, Anna and Conerly, Tom and DasSarma, Nova and Drain, Dawn and Ganguli, Deep and Hatfield-Dodds, Zac and Hernandez, Danny and Jones, Andy and Kernion, Jackson and Lovitt, Liane and Ndousse, Kamal and Amodei, Dario and Brown, Tom and Clark, Jack and Kaplan, Jared and McCandlish, Sam and Olah, Chris},
   year={2021},
   journal={Transformer Circuits Thread},
   note={https://transformer-circuits.pub/2021/framework/index.html}
}

@article{su2025concepts,
  title={From Concepts to Components: Concept-Agnostic Attention Module Discovery in Transformers},
  author={Su, Jingtong and Kempe, Julia and Ullrich, Karen},
  journal={arXiv preprint arXiv:2506.17052},
  year={2025}
}

@inproceedings{
gandelsman2024interpreting,
title={Interpreting {CLIP}'s Image Representation via Text-Based Decomposition},
author={Yossi Gandelsman and Alexei A Efros and Jacob Steinhardt},
booktitle={The Twelfth International Conference on Learning Representations},
year={2024},
url={https://openreview.net/forum?id=5Ca9sSzuDp}
}

@inproceedings{dunesky2024transcoders,
 author = {Dunefsky, Jacob and Chlenski, Philippe and Nanda, Neel},
 booktitle = {Advances in Neural Information Processing Systems},
 editor = {A. Globerson and L. Mackey and D. Belgrave and A. Fan and U. Paquet and J. Tomczak and C. Zhang},
 pages = {24375--24410},
 publisher = {Curran Associates, Inc.},
 title = {Transcoders find interpretable LLM feature circuits},
 url = {https://proceedings.neurips.cc/paper_files/paper/2024/file/2b8f4db0464cc5b6e9d5e6bea4b9f308-Paper-Conference.pdf},
 volume = {37},
 year = {2024}
}

@inproceedings{
huben2024sparse,
title={Sparse Autoencoders Find Highly Interpretable Features in Language Models},
author={Robert Huben and Hoagy Cunningham and Logan Riggs Smith and Aidan Ewart and Lee Sharkey},
booktitle={The Twelfth International Conference on Learning Representations},
year={2024},
url={https://openreview.net/forum?id=F76bwRSLeK}
}

@article{olshausen1997sparse,
  title={Sparse coding with an overcomplete basis set: A strategy employed by V1?},
  author={Olshausen, Bruno A and Field, David J},
  journal={Vision research},
  volume={37},
  number={23},
  pages={3311--3325},
  year={1997},
  publisher={Elsevier}
}

@article{lee2006efficient,
  title={Efficient sparse coding algorithms},
  author={Lee, Honglak and Battle, Alexis and Raina, Rajat and Ng, Andrew},
  journal={Advances in neural information processing systems},
  volume={19},
  year={2006}
}

@article{elhage2022superposition,
   title={Toy Models of Superposition},
   author={Elhage, Nelson and Hume, Tristan and Olsson, Catherine and Schiefer, Nicholas and Henighan, Tom and Kravec, Shauna and Hatfield-Dodds, Zac and Lasenby, Robert and Drain, Dawn and Chen, Carol and Grosse, Roger and McCandlish, Sam and Kaplan, Jared and Amodei, Dario and Wattenberg, Martin and Olah, Christopher},
   year={2022},
   journal={Transformer Circuits Thread},
   note={https://transformer-circuits.pub/2022/toy\_model/index.html}
}

@article{he2025understandingnatureattentionlowrank,
      title={Towards Understanding the Nature of Attention with Low-Rank Sparse Decomposition}, 
      author={Zhengfu He and Junxuan Wang and Rui Lin and Xuyang Ge and Wentao Shu and Qiong Tang and Junping Zhang and Xipeng Qiu},
      year={2025},
      eprint={2504.20938},
      journal={arXiv preprint arXiv:2504.20938},
      archivePrefix={arXiv},
      primaryClass={cs.LG},
      url={https://arxiv.org/abs/2504.20938}, 
}

@inproceedings{gated-sae,
 author = {Rajamanoharan, Senthooran and Conmy, Arthur and Smith, Lewis and Lieberum, Tom and Varma, Vikrant and Kram\'{a}r, J\'{a}nos and Shah, Rohin and Nanda, Neel},
 booktitle = {Advances in Neural Information Processing Systems},
 doi = {10.52202/079017-0024},
 editor = {A. Globerson and L. Mackey and D. Belgrave and A. Fan and U. Paquet and J. Tomczak and C. Zhang},
 pages = {775--818},
 publisher = {Curran Associates, Inc.},
 title = {Improving Sparse Decomposition of Language Model Activations with Gated Sparse Autoencoders},
 url = {https://proceedings.neurips.cc/paper_files/paper/2024/file/01772a8b0420baec00c4d59fe2fbace6-Paper-Conference.pdf},
 volume = {37},
 year = {2024}
}

@article{dyslexify,
  title={Towards mechanistic defenses against typographic attacks in clip},
  author={Hufe, Lorenz and Venhoff, Constantin and Dreyer, Maximilian and Lapuschkin, Sebastian and Samek, Wojciech},
  journal={arXiv preprint arXiv:2508.20570},
  year={2025}
}

@article{ilharco2022patching,
  title={Patching open-vocabulary models by interpolating weights},
  author={Ilharco, Gabriel and Wortsman, Mitchell and Gadre, Samir Yitzhak and Song, Shuran and Hajishirzi, Hannaneh and Kornblith, Simon and Farhadi, Ali and Schmidt, Ludwig},
  journal={Advances in Neural Information Processing Systems},
  volume={35},
  pages={29262--29277},
  year={2022}
}

@article{wu2025qwen,
  title={Qwen-image technical report},
  author={Wu, Chenfei and Li, Jiahao and Zhou, Jingren and Lin, Junyang and Gao, Kaiyuan and Yan, Kun and Yin, Sheng-ming and Bai, Shuai and Xu, Xiao and Chen, Yilei and others},
  journal={arXiv preprint arXiv:2508.02324},
  year={2025}
}

@article{schuhmann2022laion,
  title={Laion-5b: An open large-scale dataset for training next generation image-text models},
  author={Schuhmann, Christoph and Beaumont, Romain and Vencu, Richard and Gordon, Cade and Wightman, Ross and Cherti, Mehdi and Coombes, Theo and Katta, Aarush and Mullis, Clayton and Wortsman, Mitchell and others},
  journal={Advances in neural information processing systems},
  volume={35},
  pages={25278--25294},
  year={2022}
}

@inproceedings{denseprefix,
  title={Defense-prefix for preventing typographic attacks on clip},
  author={Azuma, Hiroki and Matsui, Yusuke},
  booktitle={Proceedings of the IEEE/CVF International Conference on Computer Vision},
  pages={3644--3653},
  year={2023}
}
\clearpage
\appendix
\crefalias{section}{appendix}
\crefalias{subsection}{appendix}

\section*{\LARGE Appendix}

\section{Broader Impact}
\label{app:impact}
This work introduces and evaluates an intervention method against typographic attacks. We recognize the potential risks in releasing the method and results, but believe the advances in mechanistic understanding of how ViT processes information from distinct modalities outweigh these risks and promote stronger disentanglement and defense. 

\section{Datasets and Text Augmentation}
\subsection{Dataset Details}
\noindent \textbf{Training Dataset.} We train our sparse transcoder on ImageNet~\citep{imnet} with the self-supervised objective. 

\noindent \textbf{Evaluation Datasets.} We evaluate the defense methods on the following datasets:
\begin{itemize}
    \item \textbf{RTA-100}~\citep{denseprefix} consists of 100 categories and a 1000-item mixture of synthetic and real-world typographic attack images. The images are drawn from 10 regular image datasets, 2 typographic attack datasets, and additional images collected from the literature. 
    \item \textbf{Disentangling}~\citep{disentangle} consists of 19 categories and 171 images, constructed with daily objects and attack text written on sticky notes. 
    \item \textbf{PAINT}~\citep{ilharco2022patching} contains 110 images of daily objects and attack-text sticky notes.
    \item \textbf{ImageNet-100-Text-D} is constructed from a 100-class subset of ImageNet~\citep{clane_im100}. We use the validation set of 5000 images for data augmentation with Qwen-Image-Edit~\citep{wu2025qwen}. We consider seven common types of texts that often coexist with objects across various scenarios and are visualized in ways not covered by previous datasets: neon sign, comic chat balloon, street sign, billboard, engraved metal plate, sculpted 3D CGI text, and obvious watermark. The prompt for generating the text-injected images is shown in \Cref{fig:qwen-prompt}.
    \item \textbf{ImageNet-100}~\citep{shekhar2021imagenet100} is a subset of ImageNet consisting of 100 classes. We use it to evaluate defense tradeoffs in regular image classification accuracy and compare with other methods. 
\end{itemize}
\begin{figure}[!htbp]
    \centering
    \begin{promptbox}[title={User Prompt}]
    <image> Add a \{text\_type\} displaying the text "\{text\_word\}" next to the \{gt\_text\}.
    \end{promptbox}
    \caption{\textbf{Prompt for Typographic Image Editing.} text\_type refers to one of the seven text types we defined. text\_word is chosen randomly from the class label set, excluding the ground truth label of the input image. gt\_text refers to the ground truth label of the input image.}
    \label{fig:qwen-prompt}
\end{figure}

\subsection{Text Injection Augmentation}
\label{sec:augment}
The text injection augmentation for nTAS calculation is defined as follows. Given an unlabeled image dataset $D$ and a reference text set $T$, for each image, we randomly choose text $t\in T$ and text position index $p \in \{0,1,2,3\}$, where $0,1,2,3$ correspond to up, down, left, and right, respectively. Then we define the shrink factor $k = 0.2$ and resize the image with original shape $(h,w)$ into $((1-k)\cdot h, w)$ if $p\in \{0,1\}$, or $(h, (1-k) \cdot w)$ if $p \in \{2,3\}$. The text is rendered at the corresponding side $p$ as defined above with shape $(k\cdot h, w)$ if $p\in \{0,1\}$, or $(h, k \cdot w)$ if $p \in \{2,3\}$ . Since we know the location of text during augmentation, it is trivial to obtain the text mask $\mathbf{m}_i$ in \maincref{eq:tas}{the nTAS definition}. 

\subsection{ImageNet-100-Text-D Samples}
\label{app:data}
\begin{figure}[!htbp]
    \centering
    \includegraphics[width=\linewidth]{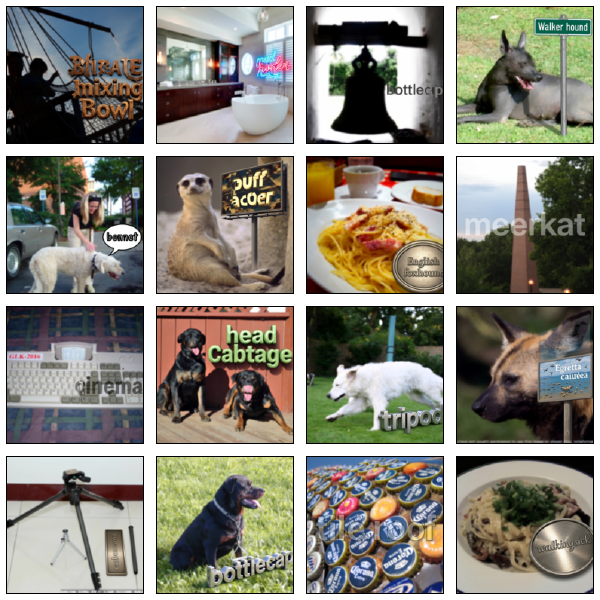}
    \caption{\textbf{Sampled Images from  ImageNet-100-Text-D.}}
    \label{fig:data_viz}
\end{figure}

\Cref{fig:data_viz} shows 16 images from the dataset. Notably, this dataset contains a broad range of text occurrences and object contexts, with occasional text-rendering errors due to an imperfect diffusion image-editing process. 

\section{Intervention Analysis on CLIP}
\subsection{Intervention Strength and Circuit Ablation Efficacy}
\label{app:sweep}
\begin{figure*}[!tp]
    \centering
    \includegraphics[width=\textwidth]{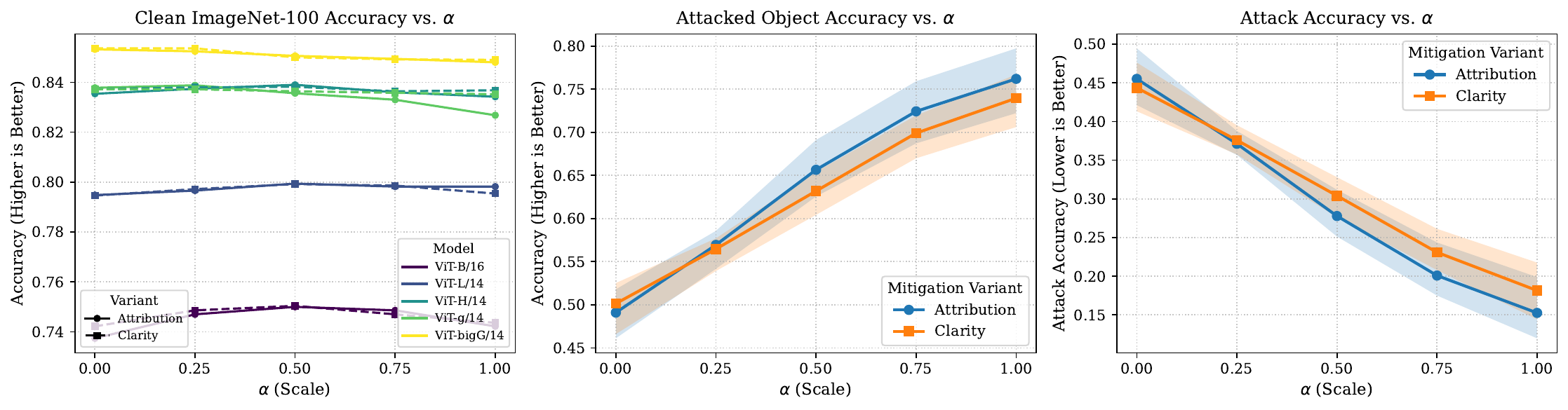}
    \caption{\textbf{Intervention Strength vs. Performance.} We sweep 5 values across [0,1] of $\alpha$, plot clean ImageNet-100 accuracy, and aggregate the Object Accuracy and the Attack Accuracy among all 5 tested CLIP models. Increasing $\alpha$ gives consistent performance gain with minimal effect on clean data accuracy.}
    \label{fig:sweep}
\end{figure*}

\Cref{fig:sweep} shows that the intervention removes the most lexical information during CLIP encoding with maximal $\alpha=1$. The sweep shows a clean progression from full lexical-circuit-induced interference at $\alpha=0 $ to strong semantic focus at $\alpha=1$. The clean ImageNet-100 test accuracy is minimally affected across values of $\alpha$. The observations support our choice of $\alpha = 1$ in the applications. 

\subsection{Clean-Data Trade-off}
\label{app:tradeoff}
As shown in the Clean column of \Cref{tab:supp_full}, our method for defense against typographic attacks minimally degrades the performance of CLIP models' zero-shot object classification, even operating with no object class labels in the clarity score ranking setup. Notably, the nTAS selection's clean-accuracy cost matches or undercuts that of supervised Defense-Prefix training on ViT-L/14, ViT-H/14, and ViT-g/14 ($-0.0$/$-0.4$/$-0.5$ vs.\ $-0.1$/$-0.4$/$-0.6$ points), while Defense-Prefix, which fine-tunes a prefix, gains clean accuracy outright on ViT-B/16 and ViT-bigG/14. Compared to \citet{dyslexify}, the nTAS selection and all four gated variants stay within their $1\%$ accuracy budget for greedy circuit finding on every backbone; the static CS selection exceeds it only on ViT-bigG/14 ($-3.3$, repaired to $-0.4$ by gating). Such an observation suggests that \citet{dyslexify}'s bottom-up greedy circuit search based on attention-weight ranking suffers from noisy module ranking and may operate on correlation rather than causality: a text patch can receive high attention weights, but the message passed by the attention head to downstream modules may not be lexical shape. By contrast, our top-down method uses sparse decompositions of the attention head outputs to interpret the internal information processing of individual attention heads, eliminating the need for incremental greedy circuit search and improving attribution faithfulness. The causal relations are shown via gradient-based attribution in \maincref{fig:viz}{the attribution visualizations}.

\begin{figure}[!htbp]
    \centering
    \subfloat[Obj: \textit{boar}, Atk: \textit{cauliflower}.\label{fig:casesim}]{\includegraphics[width=0.5\linewidth]{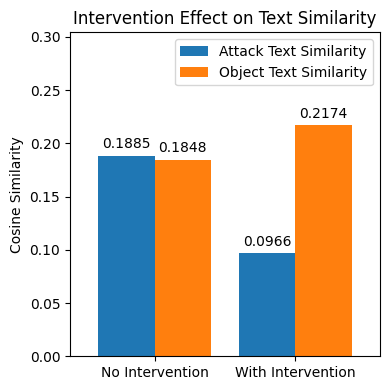}}
    \subfloat[Obj: \textit{melon}, Atk: \textit{pot}.\label{fig:rta_casesim}]{\includegraphics[width=0.5\linewidth]{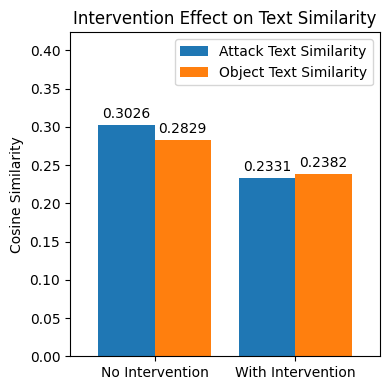}}
    \caption{\textbf{Changes in object and attack text similarity.} The bar plot shows the changes in cosine similarities of the CLIP image embedding to both the object text (Obj) and attack text (Atk). (a) The ImageNet-100-Text-D case. (b) The RTA-100 case. }
    \label{fig:casesim_all}
\end{figure}

\subsection{ViT-B/16 Case Study}
\label{sec:case}
\begin{figure*}[!tp]
    \centering
    \begin{tabular}{c|c}
    \subfloat[ImageNet-100-Text-D case: object \textit{boar}, attack
    \textit{cauliflower}, injected as a \textit{neon sign}.\label{fig:case}]{%
      \includegraphics[width=0.42\textwidth]{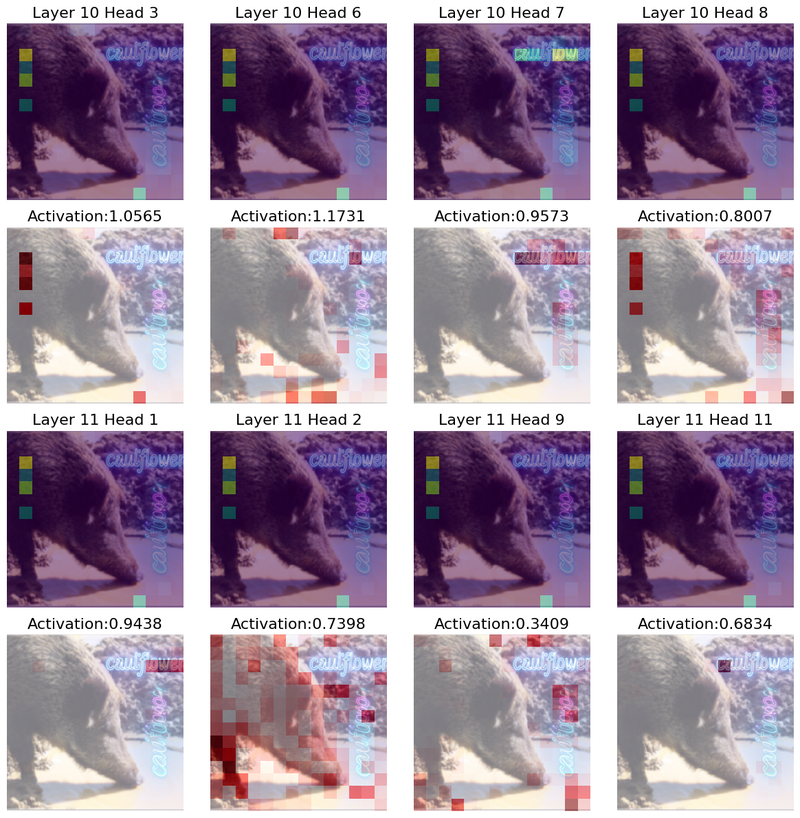}} 
         &  
    \subfloat[RTA-100 case: object \textit{melon}, attack \textit{pot},
    injected via a \textit{sticker note}.\label{fig:rta_case}]{%
      \includegraphics[width=0.42\textwidth]{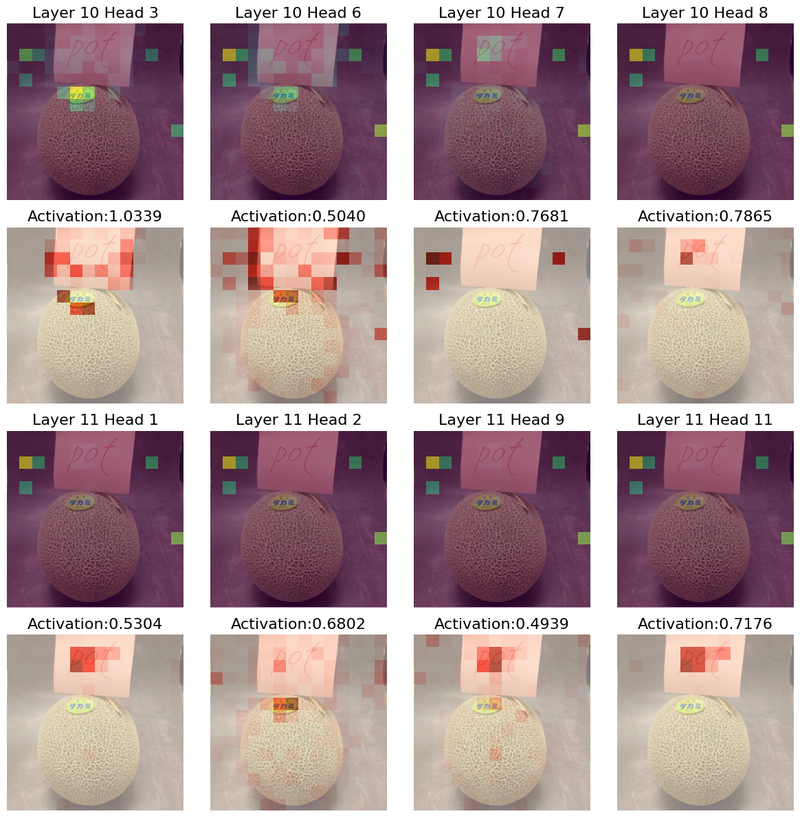}}     
    \end{tabular}
    \caption{\textbf{Case visualizations.} Each panel shows the
    text-injected image overlaid with the patch attention map of the
    <cls> token (rows 1, 3) and the average attribution maps of the
    Top-3 activations in each head (rows 2, 4). Post-intervention
    visualizations are omitted: patch tokens do not contribute to the
    <cls> representation at reweighting parameter $\alpha = 1$.}
    \label{fig:cases_pair}
\end{figure*}

This section shows our intervention in action by visualizing attention and attribution maps from in-circuit heads. The changes in the image embeddings introduced by the intervention are demonstrated by examining the cosine similarities between the image embedding and the object text, and between the image embedding and the attack text. 
We present two cases: one from our diverse synthetic ImageNet-100-Text-D dataset (Obj: \textit{boar}, Atk: \textit{cauliflower}) and the other from the simple RTA-100 dataset (Obj: \textit{melon}, Atk: \textit{pot}). 
 \Cref{fig:casesim} and \Cref{fig:rta_casesim} show that our typographic circuit mining and attention reweighting intervention has successfully corrected the error in classification in both cases.
 A more detailed inspection in \Cref{fig:case} shows that the in-circuit attention heads produce activations that are attributed to the text regions in the image. The strength of attribution over attention is demonstrated in Layer~10~Head~7 and Layer~11~Head~1, where attention maps produce distracting maps to the text, whereas attribution yields faithful maps in the text region.

The case from RTA-100 in \Cref{fig:rta_case} demonstrates further the robustness of our circuit mining method. The Layer~11~Head~11 attribution map does not fully cover the attack text in the \textit{boar}-\textit{cauliflower} example, but in the \textit{melon}-\textit{pot} example yields accurate attribution to the text. Conversely, the Layer~10~Head~7 attribution map captures the \textit{cauliflower} text successfully, but does not cover the \textit{pot} text. These changes in the behavior of the same attention head across images and texts indicate a division of labor among attention heads in processing lexical information. Our circuit mining method nonetheless captures these typographic-focusing heads despite such behavioral variance, demonstrating robustness.

\providecommand{\tablebox}[1]{\resizebox{\ifdim\width>\linewidth\linewidth\else\width\fi}{!}{#1}}

\section{Red-Teaming the Clarity Score}
\label{supp:redteam}

Scoring-corpus contamination is not an operational attack scenario: the
scoring corpus is chosen by the model owner, offline. We nevertheless
stress-tested the metrics with adversarial text corpora to
characterize their failure modes. Six conditions were applied to controlled 2{,}500 images: no overlay, one repeated word, a
5-word set, a 100-word set, scrambled glyphs, and a non-text overlay
of the same geometry.

\begin{figure}[!htbp]
\centering
\includegraphics[width=0.9\linewidth]{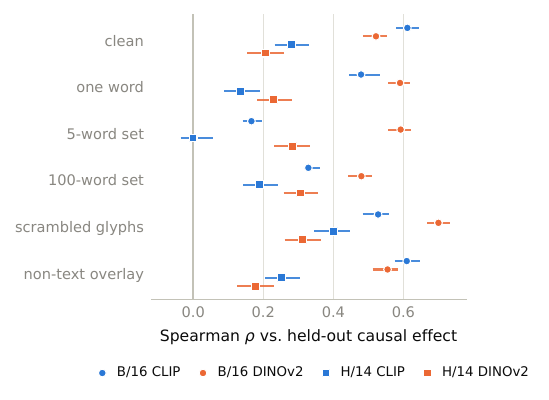}
\caption{\textbf{Red-teaming the Clarity Score.} $95\%$ bootstrap CIs: CLIP-space
scoring collapses under low-diversity text; DINOv2-space scoring of the
same top-activating sets does not.}
\label{fig:cscontrols}
\end{figure}

\paragraph{Failure mode and mechanism.}
With $100$ distinct words for circuit mining, text-firing latents
curate word-concentrated top sets: $69\%$ (B/16, layer 11) and
$49$--$78\%$ (H/14, layers 26--31) have a single word on at least half
of their top-10 images, and word concentration predicts a latent's
CLIP-space CS at $\rho = 0.80$--$0.90$ across these layers.
Consequently (\Cref{tab:cscontrols}, \Cref{fig:cscontrols}),
CS scored on low-diversity text corpora collapses and the latent-level separation of
text-firing latents under a single repeated word falls below our
criterion fixed in advance of evaluation.

\paragraph{Remedy.}
Scoring the \emph{same} top-activating sets in DINOv2 space, where no
language supervision exists to cluster shared words, remains predictive
under every manipulation: the geometric signal is genuine, and the
CLIP embedding was the wrong measuring instrument for contaminated
corpora. Note these paired-stimulus values are not comparable to the
main text's natural-corpus budget curve (a larger and easier scoring
setting; H/14 reaches $0.280$ on this corpus vs.\ $0.374$--$0.404$ on
natural corpora).

\paragraph{Reading \Cref{tab:cscontrols} and \Cref{fig:cscontrols}.}
Each cell of \Cref{tab:cscontrols} is a Spearman rank correlation
$\rho$ computed as follows: score every late-layer head's CS on the
corpus named by the row (2{,}500 held-out images, overlay content
varied only), rank heads by that score, and correlate the ranking
against the same heads' \emph{held-out causal effect} --- the
per-head object-accuracy recovery measured by suppressing that head
alone on the causal-evaluation split, which serves as ground truth for
``is this head worth muting.'' Column pairs give the embedding space
in which CS's cluster concentration is measured: CLIP space (the
metric as used in the main text) and DINOv2 space (same top-activating
image sets, re-embedded in a language-free encoder). The message is
the interaction: rows with low text diversity (one word, 5-word set)
destroy the CLIP-space columns (B/16 $0.611 \rightarrow 0.166$; H/14
to exactly $0.000$) because shared rendered words fake cluster
tightness in a language-supervised space, while the DINOv2 columns
stay in their normal range under every row --- the top-activating sets
themselves remain informative; only the measuring space was corrupted.
\Cref{fig:cscontrols} plots the same twelve values with $95\%$
bootstrap confidence intervals (resampling heads), so the collapse
cells can be distinguished from noise: the H/14 5-word CLIP-space CI
straddles zero, whereas every DINOv2 CI excludes it.

\begin{table}[!htbp]
\centering
\small
\setlength{\tabcolsep}{4.5pt}
\tablebox{%
\begin{tabular}{lrrrr}
\toprule
 & \multicolumn{2}{c}{CLIP space} & \multicolumn{2}{c}{DINOv2 space} \\
Corpus text & B/16 & H/14 & B/16 & H/14 \\
\midrule
none (clean)      & $0.611$ & $0.280$ & $0.522$ & $0.206$ \\
one word          & $0.479$ & $0.135$ & $0.590$ & $0.230$ \\
5-word set        & $0.166$ & $\mathbf{0.000}$ & $0.592$ & $0.284$ \\
100-word set      & $0.329$ & $0.190$ & $0.480$ & $0.306$ \\
scrambled glyphs  & $0.528$ & $0.401$ & $0.700$ & $0.313$ \\
non-text overlay  & $0.610$ & $0.251$ & $0.555$ & $0.177$ \\
\bottomrule
\end{tabular}}
\caption{\textbf{Red-teaming the Clarity Score.} Spearman $\rho$ between the CS head
ranking (scored on each contaminated corpus) and held-out causal
effect. In CLIP space low-diversity text destroys the signal (H/14 CI
$[-0.04, +0.06]$ contains zero); in a language-free space the same
top-activating sets remain predictive throughout.}
\label{tab:cscontrols}
\end{table}

\section{Content Sensitivity of the Text-Attribution Signal}
\label{supp:contentcontrols}

With identical border geometry per image, we compare real words against
scrambled glyphs, a texture patch, a solid rectangle, and a blank
border. Two findings. (i) \emph{nTAS detects rendered script, not
readability}: scrambled glyphs reproduce the real-text head ranking
almost exactly (rank correlation $0.978$ on B/16, $0.794$ on H/14) with
equal or better causal accuracy ($\rho = 0.622$/$0.385$ vs.\
$0.569$/$0.350$). This is consistent with the mechanism: heads detect
text \emph{form} . (ii)
\emph{Text specificity is model-dependent}: on H/14, non-text overlays
collapse the signal ($\rho \leq 0.175$ vs.\ $0.35$--$0.39$ for script),
cleanly dissociating script sensitivity from border salience; on B/16,
solid and blank borders retain $\rho \approx 0.55$ (texture $0.37$), so
smaller-model ``lexical'' heads overlap with generally border-attentive
heads.

\section{Architecture Ablation at Equal Capacity}
\label{supp:arch}

\begin{table}[!htbp]
\centering
\small
\setlength{\tabcolsep}{4.5pt}
\tablebox{%
\begin{tabular}{lrrrr}
\toprule
Selector & \multicolumn{2}{c}{$\rho$ vs. causal} &
\multicolumn{2}{c}{Group $\Delta$ (pt)} \\
 & B/16 & H/14 & B/16 & H/14 \\
\midrule
Head-dec. nTAS & $0.569$ & $0.350$ & $+12.5$ & $+15.3$ \\
Monolithic nTAS & $0.576$ & $0.340$ & $+14.4$ & $+17.6$ \\
Direct grad.\ ratio & $0.654$ & $0.403$ & n/a & n/a \\
Head-dec. CS (50k) & $\mathbf{0.712}$ & $\mathbf{0.404}$ & $+19.1$ & $+14.9$ \\
Monolithic CS & $0.494$ & $0.116$ & n/a & n/a \\
\bottomrule
\end{tabular}}
\caption{\textbf{Architecture ablation at equal capacity.} Head
ranking accuracy and group recovery on ImageNet-100-Text. Group
$\Delta$ here uses the \emph{held-out causal-evaluation split}; the
main table reports full-benchmark absolute accuracies, hence the
different magnitudes. ``n/a'': ranking-only baselines, evaluated for
selection accuracy but not run as group interventions. nTAS rows here
use the neutral-vocabulary scoring protocol
(\Cref{supp:contentcontrols}). CS row: $\rho$ at the 50k scoring
budget; its group $\Delta$ uses the clean-corpus selection.}
\label{tab:arch}
\end{table}

We train monolithic transcoders over concatenated
heads~\citep{kissane2024interpretingattentionlayeroutputs} at exactly the same budget (same total
latents, $K$, seed, data, trainer) and report the trade plainly.
In \Cref{tab:arch}, each row is a head-selection strategy: our
head-decomposed dictionaries scored by nTAS and by CS, the same
scores computed on the equal-capacity monolithic dictionary (latents
assigned to their dominant head), and a dictionary-free baseline that
ranks heads directly by their gradient attribution ratio. The
``$\rho$ vs.\ causal'' columns report the Spearman correlation
between each selector's head ranking and the per-head causal ground
truth defined in \Cref{supp:redteam} (higher = the selector orders
heads closer to their true causal importance). The ``Group $\Delta$''
columns report the end effect: the change in object accuracy (points)
on the held-out causal-evaluation split of ImageNet-100-Text when the
selector's chosen head group is suppressed together; ranking-only
baselines that were never run as group interventions are marked
n/a.
\emph{Costs of head decomposition}: the monolithic dictionary
reconstructs better everywhere (FVU $0.068$--$0.094$ vs.\
$0.088$--$0.120$ on B/16; $0.107$--$0.214$ vs.\ $0.136$--$0.258$ across
H/14 layers)---a $17$--$23\%$ relative penalty---and, for defense-only
head ranking, monolithic-derived and even dictionary-free direct
selectors are as good (\Cref{tab:arch}). \emph{What the constraint
buys}: monolithic latents are head-mixed at near-chance purity (dominant
head energy share $0.10$--$0.14$ vs.\ chance $0.06$--$0.08$), so
latent-to-head attribution is ill-posed without the constraint; and the
label-free CS selector---the best single selector here and the only one
usable with no synthetic text at all---degrades severely when built on
dominant-head-assigned monolithic latents ($0.712 \rightarrow 0.494$ on
B/16, $0.404 \rightarrow 0.116$ on H/14). Head-local structure is
load-bearing for label-free identification.

\section{Qualitative Concept Panels}
\label{supp:qualitative}

\begin{figure*}[!tp]
\centering
\includegraphics[width=0.92\textwidth]{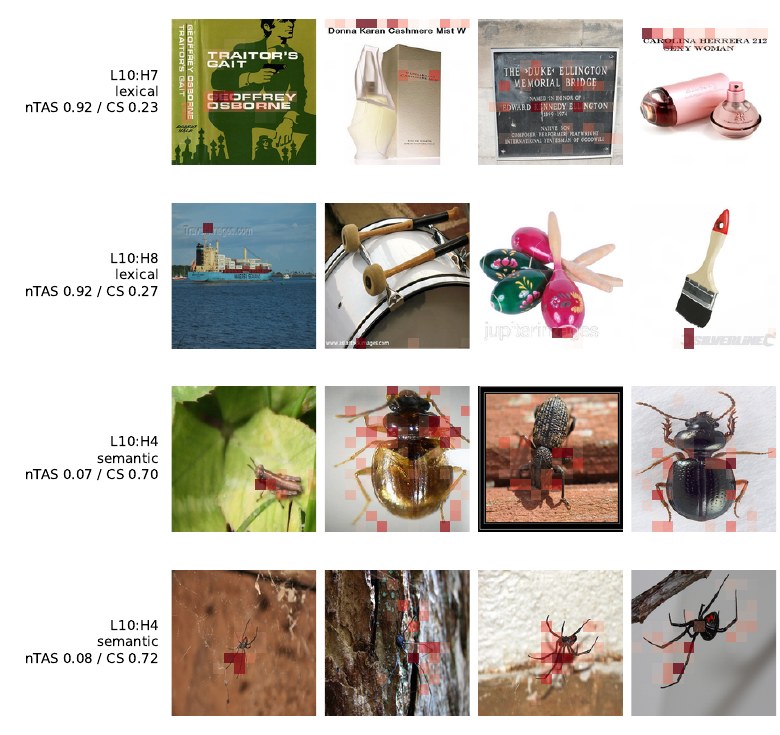}
\caption{\textbf{Concept panels, ViT-B/16.} Two lexical (top rows) and two semantic
(bottom rows) latents --- the four highest-activating images of
each latent, with its analytic attribution overlaid in red. Row labels: layer:head, population, and
nTAS / CS of the latent.}
\label{fig:supp_concepts}
\end{figure*}

\Cref{fig:supp_concepts} grounds the two latent populations in
examples. Each row is a single latent of the ViT-B/16 dictionaries:
the top two rows are the highest-nTAS latents inside the selected
lexical heads; the bottom two are low-nTAS latents drawn from the
top-CS decile of the remaining heads of the same layer (row labels
report each latent's nTAS / CS). For each latent, the four
highest-activating images on the clean corpus are shown with the
latent's analytic gradient attribution overlaid in red. The
dissociation the two scores measure is directly visible: lexical
latents fire on rendered text wherever it occurs, product labels,
book typography, stock-photo watermarks, engraved plaques, across
otherwise unrelated images (hence low CS), and their attribution
lands on the text pixels themselves; semantic latents fire on
visually coherent object clusters (hence high CS), with attribution
on the object.

\begin{figure*}[!tp]
\centering
\includegraphics[width=0.92\textwidth]{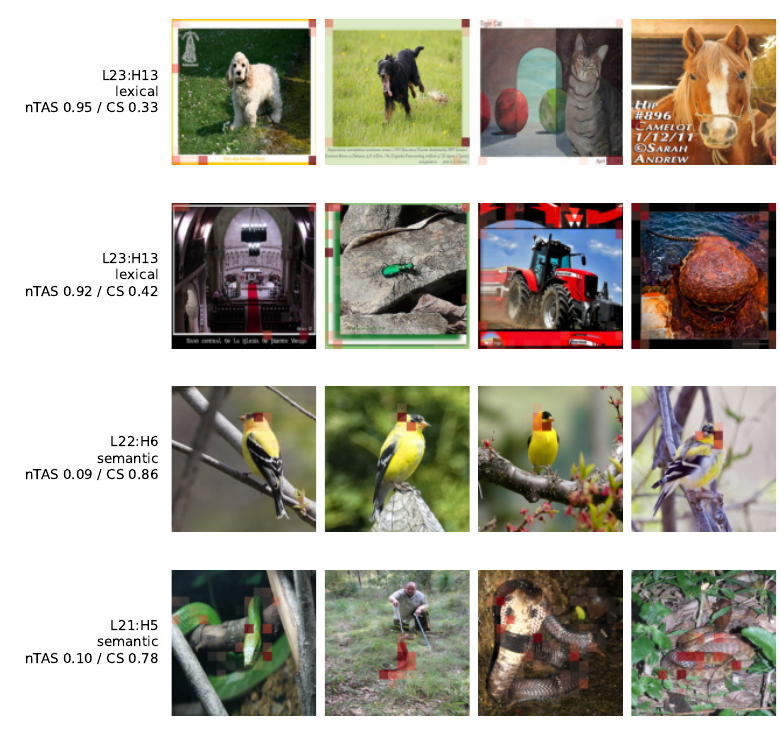}
\caption{\textbf{Concept panels, ViT-L/14.}}
\label{fig:supp_concepts_l14}
\end{figure*}

\begin{figure*}[!tp]
\centering
\includegraphics[width=0.92\textwidth]{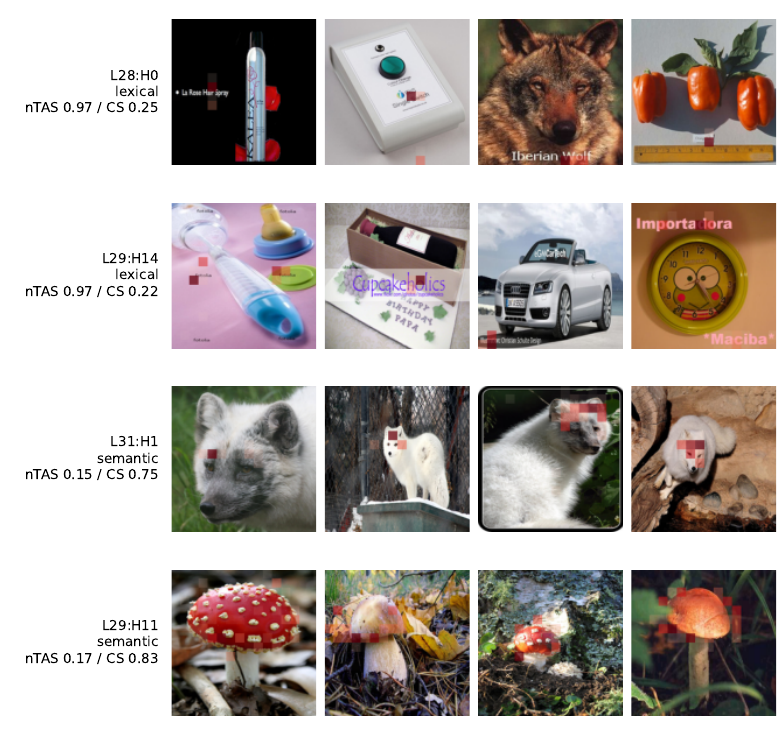}
\caption{\textbf{Concept panels, ViT-H/14.}}
\label{fig:supp_concepts_h14}
\end{figure*}

\begin{figure*}[!tp]
\centering
\includegraphics[width=0.92\textwidth]{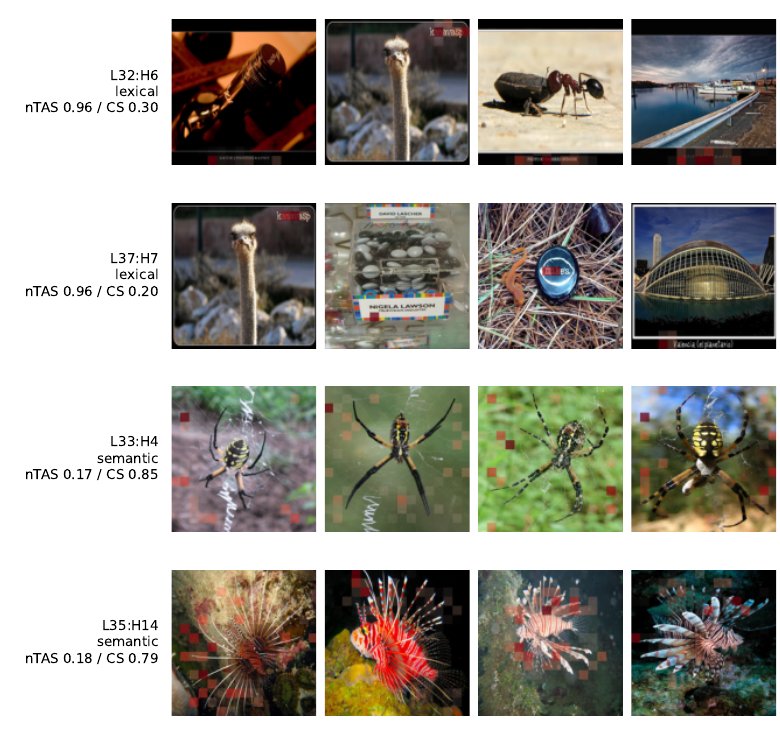}
\caption{\textbf{Concept panels, ViT-g/14.}}
\label{fig:supp_concepts_g14}
\end{figure*}

\begin{figure*}[!tp]
\centering
\includegraphics[width=0.92\textwidth]{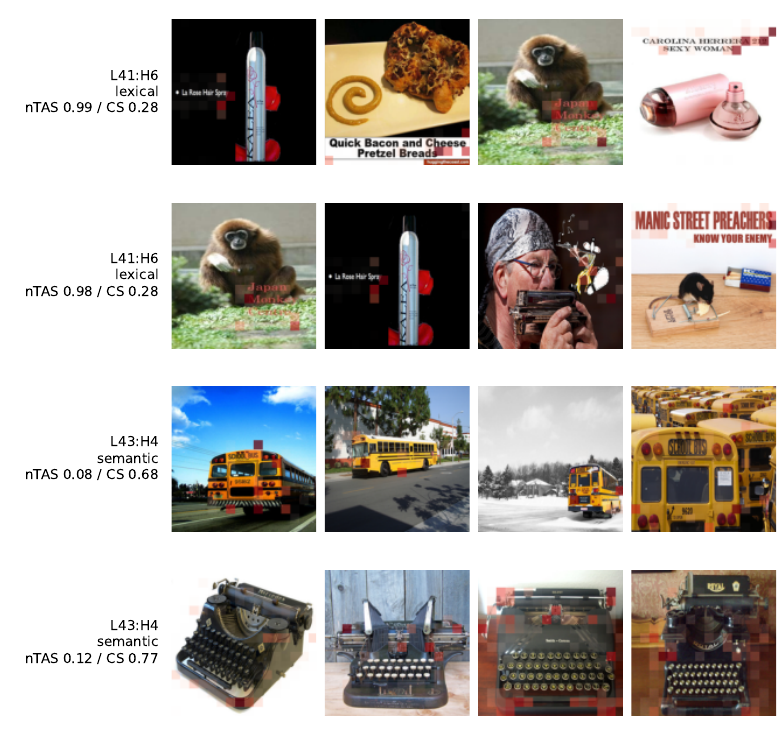}
\caption{\textbf{Concept panels, ViT-bigG/14.}}
\label{fig:supp_concepts_big}
\end{figure*}

The same construction applies unchanged to the remaining four
backbones (\Cref{fig:supp_concepts_l14,fig:supp_concepts_h14,fig:supp_concepts_g14,fig:supp_concepts_big}),
and the dissociation strengthens with scale: the selected lexical
latents' nTAS rises from $0.92$ (B/16) to $0.99$ (bigG/14) while the
semantic rows hold CS between $0.68$ and $0.86$; the lexical rows'
content ranges over product labels, captions, watermarks, posters,
and clock lettering, always with attribution on the text pixels.


\begin{figure*}[!tp]
\centering
\includegraphics[width=0.92\textwidth]{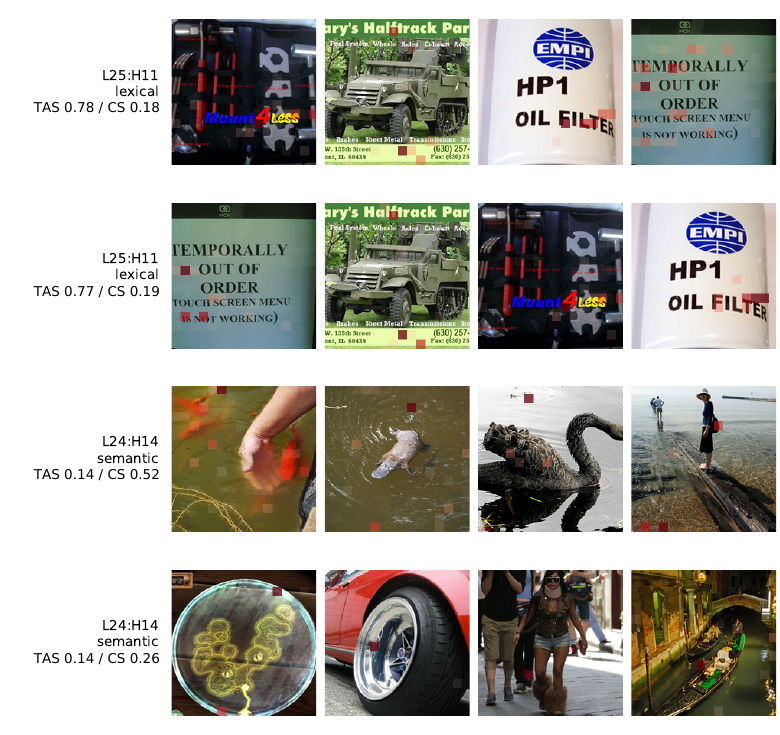}
\caption{\textbf{Concept panels, Qwen3-VL-8B vision tower.} Scores are
TAS / CS evaluated with a CLIP ViT-B/16 reference encoder.}
\label{fig:supp_concepts_qwen}
\end{figure*}

\begin{figure*}[!tp]
\centering
\includegraphics[width=0.92\textwidth]{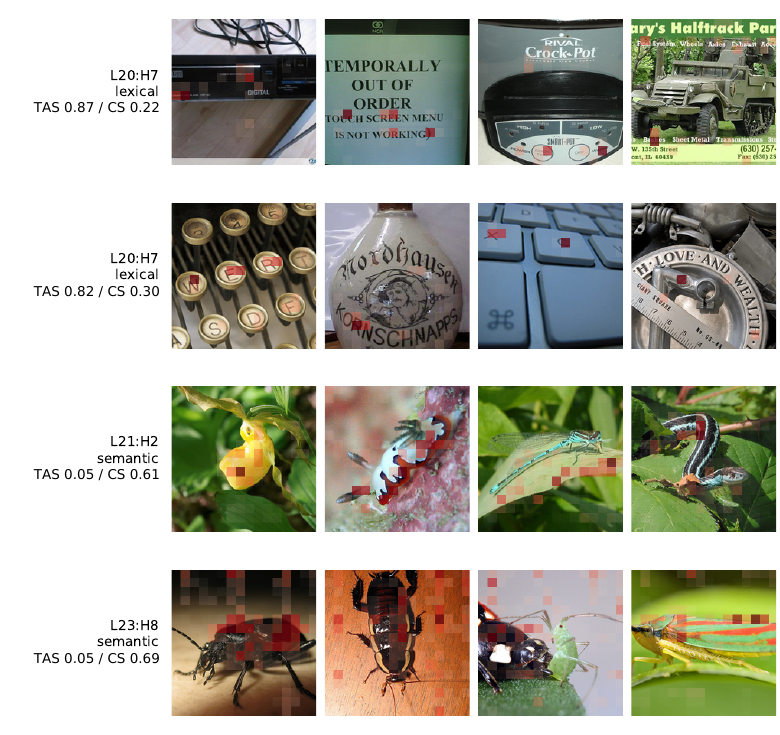}
\caption{\textbf{Concept panels, InternVL3.5-8B vision tower.} Scores are
TAS / CS evaluated with a CLIP ViT-B/16 reference encoder.}
\label{fig:supp_concepts_ivl}
\end{figure*}

\begin{figure*}[!tp]
\centering
\includegraphics[width=0.92\textwidth]{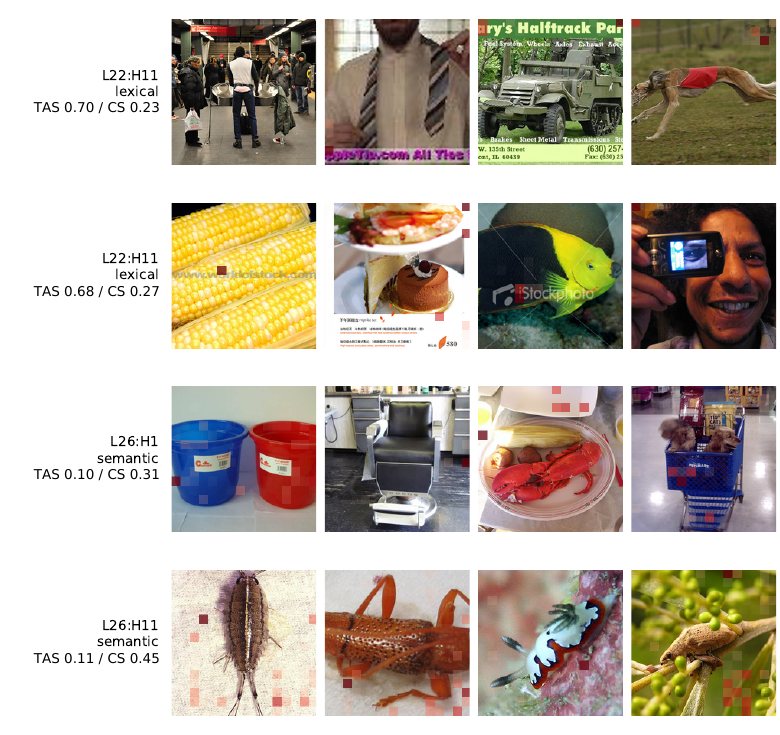}
\caption{\textbf{Concept panels, Gemma-3 12B SigLIP tower.} Scores are
TAS / CS evaluated with a CLIP ViT-B/16 reference encoder.}
\label{fig:supp_concepts_gemma}
\end{figure*}

The construction also transfers to the three LVLM vision towers
(\Cref{fig:supp_concepts_qwen,fig:supp_concepts_ivl,fig:supp_concepts_gemma}),
with two adaptations: scores are the raw text-region
attribution fraction, and clarity is computed in each tower's own
mean-pooled embedding space. Head selection uses the rendered scoring
corpus, as in the method. On all three towers, the
lexical latents' top natural images are unified by naturally
occurring text, like signage, advertisements, watermarks, menus,
product labels, with attribution on the text itself; InternVL's
second lexical latent is a particularly clean script-form detector,
firing on typewriter keys, keyboard keys, engraved bottle lettering,
and stamped metal. Semantic rows form coherent object clusters on
Qwen and InternVL, with Gemma's being the weakest.


\section{Per-Model Comparison Detail}
\label{supp:fulltable}
\Cref{tab:supp_full} expands the macro-averaged main-text comparison
to all five backbones and all methods, including the adaptive
variants. The table is method-major: each block is one intervention
strategy, with one row per CLIP backbone. Columns are
zero-shot object-classification accuracy (\%) on clean ImageNet-100
(with the signed change against that backbone's no-intervention row in
parentheses) and on the four typographic-attack benchmarks (RTA-100,
Disentangling, PAINT, ImageNet-100-Text-D). All non-starred rows are
produced by our replication run of the official code releases. The gray $^{*}$ row copies
the Dyslexify numbers as printed in the original paper, whose
evaluation protocol differs, and is shown for reference. It is
excluded from best/second-best ranking everywhere, which our methods outperform nonetheless in most cases. The
adaptive variants track their static counterparts' attack accuracy
within a few points while consistently shrinking the clean-accuracy
column's loss.

\begin{table*}[!tp]
\centering\small
\resizebox{0.95\textwidth}{!}{
\begin{tabular}{c|c|c|c|c|c|c}
\toprule
\textbf{Method} & \textbf{Model} & \textbf{IN-100 (Clean)} & \textbf{RTA-100} & \textbf{Disentangling} & \textbf{PAINT} & \textbf{IN-100-Text-D} \\
\midrule
\midrule
\multirow{5}{*}{No intervention} & ViT-B/16 & 74.8 & 56.4 & 46.2 & 59.1 & 54.6 \\
 & ViT-L/14 & 79.9 & 54.6 & 46.2 & 60.9 & 58.2 \\
 & ViT-H/14 & 83.8 & 53.4 & 39.4 & 50.9 & 56.6 \\
 & ViT-g/14 & 83.8 & 50.2 & 53.8 & 54.5 & 56.9 \\
 & ViT-bigG/14 & 85.4 & 60.9 & 42.5 & 50.0 & 62.4 \\
\midrule
\multirow{5}{*}{Defense-Prefix} & ViT-B/16 & 75.4 (+0.7) & 65.6 & 84.4 & 68.0 & 69.9 \\
 & ViT-L/14 & 79.8 (-0.1) & 62.9 & 77.2 & 71.8 & 71.6 \\
 & ViT-H/14 & 83.4 (-0.4) & 63.5 & 67.2 & 64.1 & 70.4 \\
 & ViT-g/14 & 83.2 (-0.6) & 58.1 & 60.0 & 66.0 & 66.7 \\
 & ViT-bigG/14 & 85.4 (+0.1) & 67.8 & 50.0 & 68.9 & 71.7 \\
\midrule
\multirow{5}{*}{Dyslexify} & ViT-B/16 & 75.3 (+0.5) & 67.9 & 83.9 & 69.9 & 73.5 \\
 & ViT-L/14 & 79.5 (-0.3) & 69.3 & 60.6 & 79.6 & 75.0 \\
 & ViT-H/14 & 83.4 (-0.4) & 66.1 & 53.9 & 68.9 & 70.3 \\
 & ViT-g/14 & 83.0 (-0.8) & 59.9 & 64.4 & 74.8 & 67.7 \\
 & ViT-bigG/14 & 85.1 (-0.3) & 67.5 & 55.6 & 61.2 & 70.6 \\
\midrule
\multirow{5}{*}{Dyslexify$^{*}$} & ViT-B/16 & \color{gray}75.0 (+0.2) & \color{gray}68.3 & \color{gray}85.0 & \color{gray}72.7 & -- \\
 & ViT-L/14 & \color{gray}79.5 (-0.4) & \color{gray}71.0 & \color{gray}60.6 & \color{gray}76.4 & -- \\
 & ViT-H/14 & \color{gray}83.4 (-0.4) & \color{gray}68.3 & \color{gray}72.2 & \color{gray}70.9 & -- \\
 & ViT-g/14 & \color{gray}82.6 (-1.2) & \color{gray}62.0 & \color{gray}67.2 & \color{gray}71.8 & -- \\
 & ViT-bigG/14 & \color{gray}84.7 (-0.6) & \color{gray}72.9 & \color{gray}68.3 & \color{gray}69.1 & -- \\
\midrule
\multirow{5}{*}{Ours~(CS)} & ViT-B/16 & 74.4 (-0.4) & 66.1 & 86.2 & 67.3 & 73.2 \\
 & ViT-L/14 & 79.6 (-0.3) & 66.2 & 70.0 & 67.3 & 72.8 \\
 & ViT-H/14 & 83.3 (-0.5) & 75.2 & 86.2 & 76.4 & 79.9 \\
 & ViT-g/14 & 83.0 (-0.8) & 67.2 & 86.2 & 72.7 & 75.4 \\
 & ViT-bigG/14 & 82.1 (-3.3) & 74.7 & 83.1 & 70.9 & 75.2 \\
\midrule
\multirow{5}{*}{Ours~(nTAS)} & ViT-B/16 & 74.2 (-0.5) & 68.7 & 86.9 & 70.9 & 74.2 \\
 & ViT-L/14 & 79.9 (-0.0) & 67.7 & 64.4 & 65.5 & 74.2 \\
 & ViT-H/14 & 83.4 (-0.4) & 75.6 & 79.4 & 78.2 & 79.1 \\
 & ViT-g/14 & 83.3 (-0.5) & 66.0 & 81.2 & 73.6 & 75.5 \\
 & ViT-bigG/14 & 84.7 (-0.7) & 75.8 & 66.9 & 70.0 & 80.3 \\
\midrule
\multirow{5}{*}{Ours~(CS+Adaptive)} & ViT-B/16 & 74.5 (-0.3) & 66.3 & 85.0 & 67.3 & 72.2 \\
 & ViT-L/14 & 79.8 (-0.1) & 64.7 & 67.5 & 65.5 & 73.1 \\
 & ViT-H/14 & 83.9 (+0.1) & 71.7 & 77.5 & 71.8 & 77.6 \\
 & ViT-g/14 & 83.7 (-0.1) & 65.1 & 79.4 & 71.8 & 74.5 \\
 & ViT-bigG/14 & 84.9 (-0.4) & 70.6 & 74.4 & 65.5 & 76.9 \\
\midrule
\multirow{5}{*}{Ours~(nTAS+Adaptive)} & ViT-B/16 & 74.4 (-0.3) & 67.5 & 86.2 & 70.9 & 73.8 \\
 & ViT-L/14 & 79.8 (-0.1) & 65.6 & 62.5 & 63.6 & 74.3 \\
 & ViT-H/14 & 83.6 (-0.2) & 74.8 & 77.5 & 76.4 & 78.0 \\
 & ViT-g/14 & 83.3 (-0.5) & 63.8 & 74.4 & 72.7 & 74.8 \\
 & ViT-bigG/14 & 85.0 (-0.4) & 75.2 & 64.4 & 69.1 & 79.9 \\
\bottomrule
\end{tabular}

}
\caption{\textbf{Per-model comparison.} All intervention strategies; conventions as in the main text; $^{*}$ rows as reported by the original authors (gray; their own evaluation protocol rather than ours).}
\label{tab:supp_full}
\end{table*}

\section{Statistical Detail}
\label{supp:stats}

\subsection{Main-Table Detail: Baselines, CIs, and Corrections}
\label{supp:main}

\Cref{tab:supp_main} expands the macro-averaged main-table
comparison to per-model, per-benchmark detail. Within each backbone
block, the ``Base acc.'' row gives the un-intervened zero-shot
accuracy (\%) on each benchmark column, and the four $\Delta$ rows
give the accuracy change (points) produced by Ours~(CS), Ours~(nTAS),
and their gated (+Adaptive) variants. Every $\Delta$ is computed
against \emph{that run's own} paired baseline --- intervened and
un-intervened predictions on the identical image set --- so run-level
nuisances cancel. The bracketed interval is a per-cell paired
bootstrap: images are resampled with replacement $10^4$ times, the
paired delta is recomputed on each resample, and the $2.5$th/$97.5$th
percentiles are reported. Because 20 attack cells are tested for the
CS family (five models $\times$ four attack benchmarks, one-sided),
Holm correction is applied across them: 19/20 remain significant at
$0.05$; the exception is L/14 on PAINT-DS, whose 110 images make it
the widest-CI column throughout. Two regularities to read off: every
attack-column delta is positive on every backbone for all four
variants, and the +Adaptive rows' Clean deltas never exceed $-0.5$ points in
magnitude on any backbone --- most with CIs containing zero --- with
the static outlier bigG/14 CS ($-3.3$) repaired to $-0.4$.
Random-suppression baselines are characterized distributionally in
\Cref{tab:supp_multidraw}. Selected head lists and gating thresholds
are recorded in each run's manifest.

\begin{table*}[!tp]\centering\small
\setlength{\tabcolsep}{2.6pt}
\tablebox{%
\begin{tabular}{llrrrrr}
\toprule
Model & Quantity & RTA-100 & Disent. & PAINT & IN100-T & Clean \\
\midrule
B/16 & Base acc.\ (\%) & 56.4 & 46.2 & 59.1 & 54.6 & 74.8 \\
     & CS $\Delta$ [CI] & $+9.7$ \scriptsize[$+7.5,+11.9$] & $+40.0$ \scriptsize[$+32.5,+47.5$] & $+8.2$ \scriptsize[$+0.9,+15.5$] & $+18.6$ \scriptsize[$+17.4,+19.9$] & $-0.4$ \scriptsize[$-1.0,+0.2$] \\
     & nTAS $\Delta$ [CI] & $+12.3$ \scriptsize[$+10.0,+14.6$] & $+40.6$ \scriptsize[$+33.1,+48.1$] & $+11.8$ \scriptsize[$+3.6,+20.0$] & $+19.6$ \scriptsize[$+18.4,+20.9$] & $-0.5$ \scriptsize[$-1.2,+0.1$] \\
     & CS+Adaptive $\Delta$ [CI] & $+9.9$ \scriptsize[$+7.8,+12.0$] & $+38.8$ \scriptsize[$+31.2,+46.2$] & $+8.2$ \scriptsize[$+1.8,+15.5$] & $+17.6$ \scriptsize[$+16.4,+18.9$] & $-0.3$ \scriptsize[$-0.6,+0.0$] \\
     & nTAS+Adaptive $\Delta$ [CI] & $+11.1$ \scriptsize[$+8.9,+13.4$] & $+40.0$ \scriptsize[$+32.5,+47.5$] & $+11.8$ \scriptsize[$+4.5,+19.1$] & $+19.2$ \scriptsize[$+18.0,+20.5$] & $-0.3$ \scriptsize[$-0.8,+0.2$] \\
\midrule
L/14 & Base acc.\ (\%) & 54.6 & 46.2 & 60.9 & 58.2 & 79.9 \\
     & CS $\Delta$ [CI] & $+11.6$ \scriptsize[$+9.3,+14.0$] & $+23.8$ \scriptsize[$+16.9,+30.6$] & $+6.4$ \scriptsize[$-2.7,+15.5$] & $+14.6$ \scriptsize[$+13.5,+15.7$] & $-0.3$ \scriptsize[$-0.9,+0.3$] \\
     & nTAS $\Delta$ [CI] & $+13.1$ \scriptsize[$+10.7,+15.5$] & $+18.1$ \scriptsize[$+11.9,+24.4$] & $+4.5$ \scriptsize[$-4.5,+13.6$] & $+16.0$ \scriptsize[$+14.9,+17.2$] & $-0.0$ \scriptsize[$-0.6,+0.6$] \\
     & CS+Adaptive $\Delta$ [CI] & $+10.1$ \scriptsize[$+8.1,+12.2$] & $+21.2$ \scriptsize[$+15.0,+27.5$] & $+4.5$ \scriptsize[$-3.6,+12.7$] & $+14.9$ \scriptsize[$+13.8,+16.0$] & $-0.1$ \scriptsize[$-0.3,+0.2$] \\
     & nTAS+Adaptive $\Delta$ [CI] & $+11.0$ \scriptsize[$+8.8,+13.2$] & $+16.2$ \scriptsize[$+10.0,+22.5$] & $+2.7$ \scriptsize[$-5.5,+10.9$] & $+16.1$ \scriptsize[$+15.0,+17.3$] & $-0.1$ \scriptsize[$-0.4,+0.2$] \\
\midrule
H/14 & Base acc.\ (\%) & 53.4 & 39.4 & 50.9 & 56.6 & 83.8 \\
     & CS $\Delta$ [CI] & $+21.8$ \scriptsize[$+19.0,+24.6$] & $+46.9$ \scriptsize[$+39.4,+54.4$] & $+25.5$ \scriptsize[$+16.4,+34.5$] & $+23.3$ \scriptsize[$+22.0,+24.6$] & $-0.5$ \scriptsize[$-1.0,+0.1$] \\
     & nTAS $\Delta$ [CI] & $+22.2$ \scriptsize[$+19.4,+25.0$] & $+40.0$ \scriptsize[$+31.9,+48.1$] & $+27.3$ \scriptsize[$+18.2,+36.4$] & $+22.5$ \scriptsize[$+21.2,+23.7$] & $-0.4$ \scriptsize[$-0.9,+0.0$] \\
     & CS+Adaptive $\Delta$ [CI] & $+18.3$ \scriptsize[$+15.7,+20.9$] & $+38.1$ \scriptsize[$+30.6,+45.6$] & $+20.9$ \scriptsize[$+11.8,+30.0$] & $+20.9$ \scriptsize[$+19.7,+22.1$] & $+0.1$ \scriptsize[$-0.2,+0.5$] \\
     & nTAS+Adaptive $\Delta$ [CI] & $+21.4$ \scriptsize[$+18.7,+24.1$] & $+38.1$ \scriptsize[$+30.6,+46.2$] & $+25.5$ \scriptsize[$+17.3,+33.6$] & $+21.4$ \scriptsize[$+20.2,+22.6$] & $-0.2$ \scriptsize[$-0.5,+0.2$] \\
\midrule
g/14 & Base acc.\ (\%) & 50.2 & 53.8 & 54.5 & 56.9 & 83.8 \\
     & CS $\Delta$ [CI] & $+17.0$ \scriptsize[$+14.5,+19.5$] & $+32.5$ \scriptsize[$+25.6,+40.0$] & $+18.2$ \scriptsize[$+9.1,+27.3$] & $+18.5$ \scriptsize[$+17.4,+19.7$] & $-0.8$ \scriptsize[$-1.3,-0.2$] \\
     & nTAS $\Delta$ [CI] & $+15.7$ \scriptsize[$+13.3,+18.2$] & $+27.5$ \scriptsize[$+20.6,+34.4$] & $+19.1$ \scriptsize[$+10.0,+28.2$] & $+18.6$ \scriptsize[$+17.4,+19.8$] & $-0.5$ \scriptsize[$-1.1,+0.1$] \\
     & CS+Adaptive $\Delta$ [CI] & $+14.8$ \scriptsize[$+12.5,+17.1$] & $+25.6$ \scriptsize[$+18.8,+32.5$] & $+17.3$ \scriptsize[$+10.0,+25.5$] & $+17.6$ \scriptsize[$+16.5,+18.7$] & $-0.1$ \scriptsize[$-0.5,+0.3$] \\
     & nTAS+Adaptive $\Delta$ [CI] & $+13.5$ \scriptsize[$+11.2,+15.8$] & $+20.6$ \scriptsize[$+13.8,+27.5$] & $+18.2$ \scriptsize[$+10.0,+26.4$] & $+17.9$ \scriptsize[$+16.8,+19.0$] & $-0.5$ \scriptsize[$-0.9,-0.1$] \\
\midrule
bigG/14 & Base acc.\ (\%) & 60.9 & 42.5 & 50.0 & 62.4 & 85.4 \\
     & CS $\Delta$ [CI] & $+13.8$ \scriptsize[$+11.2,+16.5$] & $+40.6$ \scriptsize[$+33.1,+48.1$] & $+20.9$ \scriptsize[$+11.8,+30.0$] & $+12.8$ \scriptsize[$+11.6,+14.1$] & $-3.3$ \scriptsize[$-4.0,-2.6$] \\
     & nTAS $\Delta$ [CI] & $+14.8$ \scriptsize[$+12.5,+17.2$] & $+24.4$ \scriptsize[$+18.1,+31.2$] & $+20.0$ \scriptsize[$+12.7,+28.2$] & $+18.0$ \scriptsize[$+16.9,+19.2$] & $-0.6$ \scriptsize[$-1.1,-0.2$] \\
     & CS+Adaptive $\Delta$ [CI] & $+9.6$ \scriptsize[$+7.4,+11.7$] & $+31.9$ \scriptsize[$+25.0,+39.4$] & $+15.5$ \scriptsize[$+7.3,+23.6$] & $+14.6$ \scriptsize[$+13.4,+15.7$] & $-0.4$ \scriptsize[$-0.7,-0.1$] \\
     & nTAS+Adaptive $\Delta$ [CI] & $+14.2$ \scriptsize[$+12.0,+16.5$] & $+21.9$ \scriptsize[$+15.6,+28.7$] & $+19.1$ \scriptsize[$+11.8,+27.3$] & $+17.6$ \scriptsize[$+16.4,+18.7$] & $-0.4$ \scriptsize[$-0.7,-0.0$] \\
\bottomrule
\end{tabular}}
\caption{\textbf{Main-table detail.} Baseline accuracies and per-cell paired-bootstrap deltas with $95\%$ CIs ($10^4$ resamples) for both selection signals and their gated variants, computed against each run's own paired baseline. Holm correction over the 20-cell CS attack family (five models $\times$ four attack sets, one-sided): 19/20 cells remain significant at $0.05$. Random-suppression distributions: Table~\ref{tab:supp_multidraw}.}
\label{tab:supp_main}\end{table*}

\begin{table}[!htbp]\centering\small
\tablebox{%
\begin{tabular}{rrrrrr}
\toprule
$N$ & \multicolumn{2}{c}{B/16} & \multicolumn{2}{c}{H/14} & Stab.\ (min) \\
 & $\rho$ & p@$k$ & $\rho$ & p@$k$ & \\
\midrule
1k & 0.553 & 0.72 & 0.374 & 0.39 & 0.903 \\
2k & 0.607 & 0.83 & 0.370 & 0.39 & 0.956 \\
5k & 0.658 & 0.83 & 0.387 & 0.43 & 0.978 \\
10k & 0.670 & 0.83 & 0.378 & 0.43 & 0.986 \\
25k & 0.711 & 0.83 & 0.391 & 0.47 & 0.995 \\
50k & 0.712 & 0.83 & 0.404 & 0.46 & -- \\
\bottomrule\end{tabular}}
\caption{\textbf{Data-budget sweep.} Clean ImageNet-1k val (3 subsample seeds per budget; mean $\rho$ and precision@$k$; minimum pairwise cross-seed rank stability across both models).}
\label{tab:supp_budget}\end{table}

\begin{figure*}[!tp]
\centering
\subfloat[B/16]{\includegraphics[width=0.19\textwidth]{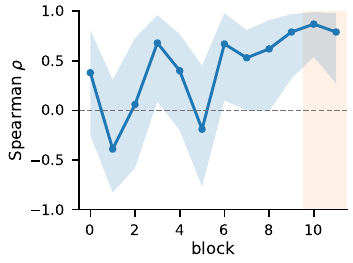}}\hfill
\subfloat[L/14]{\includegraphics[width=0.19\textwidth]{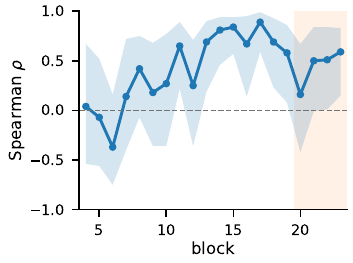}}\hfill
\subfloat[H/14]{\includegraphics[width=0.19\textwidth]{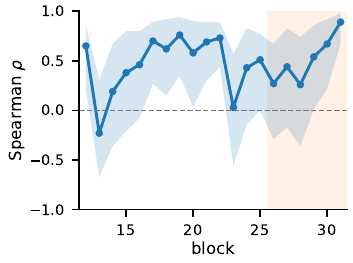}}\hfill
\subfloat[g/14]{\includegraphics[width=0.19\textwidth]{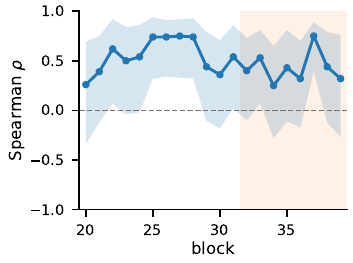}}\hfill
\subfloat[bigG/14]{\includegraphics[width=0.19\textwidth]{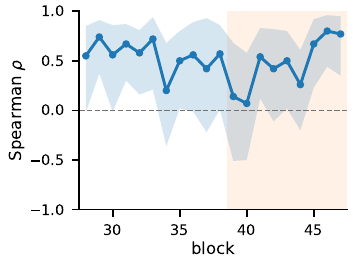}}
\caption{\textbf{Per-layer CS--nTAS agreement.} Spearman correlation at every scored
transformer block, one panel per backbone. Each point is the correlation, across
the $16$ heads of that block, between the two per-head lexicality
scores; the shaded band is the $95\%$ bootstrap CI over heads, the
dashed line marks $\rho = 0$, and the orange region is the last
$20\%$ of depth --- the region interventions draw from, and the
29-block family of the sign test.}
\label{fig:supp_layers}
\end{figure*}

\subsection{Data-Budget Sweep}
\Cref{tab:supp_budget} expands the budget curve summarized in the
main text and answers ``how much unlabeled data does CS scoring
need.'' $N$ is the scoring-corpus size, subsampled from clean
ImageNet-1k val with 3 random seeds per budget. For each seed the CS
head ranking is recomputed from scratch at that budget; the $\rho$
columns report the seed-mean Spearman correlation against the same
held-out per-head causal ground truth used throughout
(\Cref{supp:redteam}), separately for B/16 and H/14. p@$k$ is
precision-at-$k$: the fraction of the causal top-$k$ heads recovered
in the CS top-$k$, with $k$ set to the top quartile of that model's
scored heads. ``Stab.\ (min)'' is the
worst-case reproducibility number: the minimum, over both models and
all seed pairs at that budget, of the Spearman correlation between
two seeds' rankings. The message: the ranking is essentially formed
by $N{=}1$k--$2$k (stability already $\geq 0.90$), improves modestly
to a plateau at $N{=}25$k--$50$k, and the main-text operating point
($50$k) sits on that plateau rather than on a rising slope.

\subsection{Per-Layer CS--nTAS Correlations}
\Cref{fig:supp_layers} is the agreement check between our two
selection signals, resolved at every scored layer. Its purpose: CS
(geometric, computed from top-activating image clusters) and nTAS
(attributional, computed from gradients into rendered-text regions)
are derived from entirely different measurements, so if both track the
same underlying property --- how lexical a head is --- their per-head
scores should correlate wherever lexical heads exist. Each panel is
one backbone; each point gives, at that transformer block, the
Spearman correlation across the block's 16 attention heads between
their CS-derived lexicality scores and their nTAS scores on the
common axes. The shaded band is a $95\%$ CI bootstrapped over the 16
heads and is accordingly wide --- with $n{=}16$ per block, individual
blocks are not separately significant, and no per-block claim is
made. Curves start where scoring starts (for the deeper models,
mid-depth, where the SDL dictionaries are trained; e.g.\ L/14 at
block 4, bigG/14 at block 28). The orange span is the last $20\%$ of
each model's depth --- the region the intervention actually draws
from --- and its 29 blocks across the five panels form the family of
the aggregate test: all 29 sit above the dashed zero line (one-sided
sign test $p = 1.9\times10^{-9}$, stated with the caveat that blocks
within a model are not independent samples). The pattern to see is
the depth trend: early blocks fluctuate around zero (B/16 blocks
0--2, L/14 blocks 4--10), and the curves become consistently positive
from mid-depth onward, which is exactly where lexical processing
concentrates.

\subsection{Random Head Suppression over Twenty Draws}
\label{supp:multidraw}
\Cref{tab:supp_multidraw} characterizes random head suppression at the
selection's own set size: for each backbone, 20 head sets of exactly the CS
selection's size are drawn uniformly from the same late-layer pool
(seeds 1000--1019) and each is run through the identical intervention
and evaluation pipeline. Columns report, over the 20 draws, the mean
$\pm$ standard deviation and the [min, max] range of the change in
object accuracy (points) relative to no intervention --- left pair on
ImageNet-100-Text (higher is better), right pair on clean
ImageNet-100 (closer to zero is better).

\begin{table}[!htbp]
\centering
\small
\setlength{\tabcolsep}{3.2pt}
\tablebox{%
\begin{tabular}{lrrrr}
\toprule
 & \multicolumn{2}{c}{IN100-T $\Delta$ (pts)} & \multicolumn{2}{c}{Clean $\Delta$ (pts)} \\
Model & mean$\pm$sd & [min, max] & mean$\pm$sd & [min, max] \\
\midrule
B/16 & $-5.1\pm12.4$ & $[-35.9, +11.8]$ & $-11.6\pm15.1$ & $[-62.1, +0.1]$ \\
L/14 & $-0.8\pm12.7$ & $[-45.3, +11.7]$ & $-8.2\pm13.0$ & $[-55.5, -0.6]$ \\
H/14 & $+7.9\pm7.0$  & $[-14.1, +16.5]$ & $-4.3\pm4.6$  & $[-23.3, -1.5]$ \\
g/14 & $+4.8\pm9.5$  & $[-19.5, +16.4]$ & $-3.9\pm5.4$  & $[-24.0, +0.2]$ \\
bigG & $-3.7\pm17.0$ & $[-45.6, +12.6]$ & $-8.9\pm13.8$ & $[-54.3, -0.6]$ \\
\bottomrule
\end{tabular}}
\caption{\textbf{Random head suppression over twenty draws.} Head sets of the CS selection's size; change in
object accuracy versus no intervention on ImageNet-100-Text and on
clean ImageNet-100 (points).}
\label{tab:supp_multidraw}
\end{table}

Random suppression is high-variance (per-model sd $7.0$--$17.0$ points
on ImageNet-100-Text) and unreliable: mean recovery never exceeds
$+7.9$ points and is negative on three of the five models. The CS
selection's per-model recovery on ImageNet-100-Text-D ($+12.8$ to
$+23.3$; \Cref{tab:supp_full}) exceeds the
\emph{best} of the twenty random draws on every model
($+11.8$/$+11.7$/$+16.5$/$+16.4$/$+12.6$), though we note the bigG/14
margin is only $0.2$ points; the reliable contrast there is the mean
($+12.8$ vs.\ $-3.7$) and the clean cost. Random draws are also far
costlier on clean data (mean $-3.9$ to $-11.6$ points; worst draws
$-23$ to $-62$), versus at most $-3.3$ points for CS selection.

\subsection{Hybrid Selection}
Averaging the two normalized axes matches the best single signal on B/16
($+18.6$ vs.\ $+19.1$) and is best on H/14 under the evaluation-split
scoring protocol ($+16.1$, lowest clean cost); under the smaller
10k-corpus scoring protocol the H/14 hybrid degrades. We therefore
recommend the hybrid only when both signals are computed at adequate
budget, and CS alone when only label-free scoring is available.

\section{LVLM Experiments: Native-Encoder Protocol and Selection Stability}
\label{supp:lvlm}

\subsection{Protocol}
For each LVLM (Gemma-3, Qwen3-VL, InternVL3.5
families), we train Top-$K$ ($K{=}128$, seed 123, expansion $16\times$)
head-decomposed transcoders on activations of the model's \emph{native}
vision encoder (SigLIP for Gemma-3, the packed Qwen3-VL encoder,
InternViT for InternVL3.5) cached over ImageNet; heads are scored with
the same text-region attribution protocol on RIO-Bench renderings, the
selected tail is zero-ablated, and VQA accuracy is measured on
RIO-Bench splits (3{,}166 examples each). No OpenCLIP component is
transferred at any point. Head selections are stable under
independent dictionary retraining (\Cref{supp:seedstab}).

\subsection{VQA Accuracy Across Model Scales}
\label{app:vqa}

\begin{table*}[!tp]
\centering
\small
\begin{tabular}{lcccccccc}
\toprule
Model & \multicolumn{2}{c}{Clean} & \multicolumn{2}{c}{Easy} & \multicolumn{2}{c}{Medium} & \multicolumn{2}{c}{Hard} \\
\cmidrule(lr){2-3} \cmidrule(lr){4-5} \cmidrule(lr){6-7} \cmidrule(lr){8-9}
 & Base & Ours & Base & Ours & Base & Ours & Base & Ours \\
\midrule
Gemma-3 4B & 93.7 & \textbf{94.4} & 48.0 & \textbf{49.8} & 46.5 & \textbf{47.6} & 38.6 & \textbf{40.7} \\
Gemma-3 12B & 94.6 & 94.6 & 49.9 & \textbf{52.3} & 48.7 & \textbf{49.8} & 46.1 & \textbf{47.5} \\
InternVL3.5 4B & \textbf{92.0} & 91.6 & \textbf{55.1} & 54.8 & 47.4 & \textbf{47.7} & \textbf{36.1} & 35.9 \\
InternVL3.5 8B & 95.3 & 95.3 & 63.6 & \textbf{63.7} & 60.7 & 60.7 & \textbf{50.3} & 50.2 \\
InternVL3.5 14B & \textbf{92.0} & 91.8 & 56.9 & \textbf{57.0} & 55.7 & \textbf{55.8} & 46.6 & \textbf{46.7} \\
Qwen3-VL 8B & 94.9 & \textbf{95.2} & 74.1 & \textbf{75.8} & 71.3 & \textbf{73.5} & 64.5 & \textbf{66.1} \\
Qwen3-VL 30B-A3B & 83.7 & \textbf{84.1} & 67.3 & \textbf{69.6} & 66.7 & \textbf{67.8} & 60.9 & \textbf{63.1} \\
\bottomrule
\end{tabular}
\caption{\textbf{VQA accuracy under typographic attacks.} Accuracy in \%. \textbf{Base}: vision encoder without intervention; \textbf{Ours}: vision encoder with intervention. Bold marks the higher value within each pair. Clean, Easy, Medium, Hard denote the splits, with Clean split free of lexical interference.}
\label{tab:typo3_vqa_acc_pairs}
\end{table*}

To demonstrate the extension of the disentanglement to LVLMs, we test Gemma3~\citep{gemmateam2025gemma3technicalreport}, Qwen3-VL~\citep{bai2025qwen3} and InternVL3.5~\citep{wang2025internvl3} on a typographic attack VQA dataset, RIO-Bench~\citep{waseda2025read}. \Cref{tab:typo3_vqa_acc_pairs} reports VQA accuracy across the four multiple-choice splits: clean, easy, medium, and hard. Each split contains 3166 VQA triplets. We adapt the zero-ablation intervention on the lexical circuits by overriding the lexical attention head's output to zero vectors, as LVLM vision encoders may not include a dedicated class token, and all patch tokens are passed to the downstream language model.

Notably, our disentanglement and intervention pipeline increased VQA accuracy on two of the three LVLM lines tested, namely Gemma3 and Qwen3-VL, with modest margins in which head indentification matters mainly on the hardest split; we make no head-selection-superiority claim at LVLM scale. The InternVL3.5 models do not show significant improvement under the intervention, which may be due to their specific adapter design and handling of visual tokens. The three 8--12B towers of this sweep are also evaluated in \Cref{tab:lvlm_adaptive}; those counts come from a later independent execution and differ from this sweep's percentages within run-to-run evaluation noise (${\leq}0.8$ points).

\subsection{Random-Head Comparison}
A per-layer random head
specification of the same size (single draw, seed 123) evaluated identically
(\Cref{tab:supp_lvlm_rand}; Accuracy = \% of the 3{,}166
multiple-choice examples answered correctly under Base / our
selection / the random draw, ASR = attack-success rate, the \% of
attacked examples where the model outputs the attack-injected label
instead of the object --- lower is better, undefined on the clean
split) shows that on the easy and medium
splits, random same-layer ablation recovers essentially the same
accuracy as the selected heads (differences of $-5$ to $+1$ correct out
of 3{,}166, within run-to-run evaluation noise). Head \emph{identity} separates from random only on the hard
split ($+28$ correct; the random draw \emph{raises} attack-success
rate over Base by $+0.57$ points where ours lowers it) and in clean
degradation ($-0.13$ vs.\ $-0.44$ points). The Base-to-Ours deltas on
this line are small ($\approx +1$ point), and we accordingly make no
LVLM head-selection-superiority claim: the LVLM evidence establishes
that the model-specific procedure runs natively and that its selections
are reproducible, with a modest effect in which head identity matters
mainly on the hardest split. Single-draw caveat: one random
specification per split.

\begin{table}[!htbp]
\centering\small
\tablebox{%
\begin{tabular}{lrrrrrr}
\toprule
 & \multicolumn{3}{c}{Accuracy (\%)} & \multicolumn{3}{c}{ASR (\%)} \\
Split & Base & Ours & Rand. & Base & Ours & Rand. \\
\midrule
clean  & 94.76 & 94.63 & 94.31 & -- & -- & -- \\
easy   & 50.69 & 51.67 & 51.83 & 48.42 & 47.41 & 47.28 \\
medium & 48.61 & 49.65 & 49.62 & 50.66 & 49.40 & 49.53 \\
hard   & 46.65 & 46.90 & 46.02 & 52.65 & 52.40 & 53.22 \\
\bottomrule
\end{tabular}}
\caption{\textbf{Selected heads vs.\ an equal-size random draw.} Gemma-3 12B.
Base columns are identical between the two runs (bit-stable).}
\label{tab:supp_lvlm_rand}
\end{table}

\subsection{Adaptive Gating on the LVLM Towers}
We port the adaptive intervention to all three towers (firing
statistic: mean token-wise output-contribution norm; calibration on $500$
text-free ImageNet images; $\theta$ selected on a held-in half of the easy
split). Notably, tuning selects $\theta^{*}{=}0.25$ on all
three towers. \Cref{tab:lvlm_adaptive} reports correct counts ($/3166$) per
RIO-Bench split (clean, and the easy/medium/hard attack tiers) for
three variants of each tower: Base (no intervention), Static
(unconditional zero-ablation of the selected heads on every input),
and Adaptive (the same ablation applied per-image only to heads whose
firing statistic exceeds its clean calibration by $\theta^{*}$
standard deviations).
The three lines span the full regime of the mechanism, and paired
exact tests (McNemar over items, two-sided) calibrate every
comparison. On Gemma, the most typographically sensitive tower, the
gated intervention separates decisively from no intervention (easy
$p=1.6\times10^{-3}$; attack splits pooled $p=7.6\times10^{-5}$) ---
a stronger separation than static ablation achieves ($p=0.06$/$0.05$/
n.s.\ per split) --- and its point estimates exceed static on every
attack split; the adaptive--static differences themselves, however,
are within paired noise (pooled $p=0.20$), so the supported claim is
that gating matches static protection at half the (already small)
clean cost, not that it exceeds it. Head identity separates from the
equal-size random draw on the hard split ($p=0.03$; pooled
$p=0.05$). On Qwen, whose tower is more robust at baseline, the
intervention effect is the most decisive of the three lines (adaptive
vs.\ base pooled $p<10^{-13}$); static ablation is free on clean data
there, and its small attack-split advantage over gating is real
(pooled $p=3\times10^{-3}$), so on that tower always-on suppression
is simply the better operating point. On InternVL, neither static nor
adaptive ablation moves accuracy (within $\pm 3$ of base everywhere)
although the gate demonstrably fires (${\sim}1.8$ of $9$ selected
heads per image): the selected heads are not causally implicated on
this tower, consistent with its higher baseline robustness. 

\begin{table}[!htbp]
\centering
\small
\tablebox{%
\begin{tabular}{ll|cccc}
\toprule
\textbf{Tower} & \textbf{Variant} & \textbf{clean} & \textbf{easy} & \textbf{medium} & \textbf{hard} \\
\midrule
\multirow{3}{*}{Gemma-3 12B} & Base & 3000 & 1605 & 1539 & 1477 \\
 & Static & 2996 & 1636 & 1572 & 1485 \\
 & Adaptive & 2998 & \textbf{1653} & \textbf{1577} & \textbf{1495} \\
\midrule
\multirow{3}{*}{Qwen3-VL 8B} & Base & 3010 & 2349 & 2255 & 2055 \\
 & Static & 3011 & \textbf{2416} & \textbf{2319} & \textbf{2109} \\
 & Adaptive & 3010 & 2406 & 2306 & 2090 \\
\midrule
\multirow{3}{*}{InternVL3.5 8B} & Base & 3019 & 2022 & 1924 & 1586 \\
 & Static & 3017 & 2019 & 1925 & 1584 \\
 & Adaptive & 3020 & 2023 & 1923 & 1586 \\
\bottomrule
\end{tabular}}
\caption{\textbf{Adaptive gating on three LVLM vision towers.} RIO-Bench, correct
counts out of $3166$; $\theta^{*}{=}0.25$ on every tower). Bold marks the
best intervention per attack split where any intervention beats base.}
\label{tab:lvlm_adaptive}
\end{table}

\section{Random-Seed Stability of Head Selection}
\label{supp:seedstab}

Head selection depends on trained dictionaries, so we check that it is
a property of the encoder rather than of one training run's
randomness. On CLIP ViT-B/16, the late-layer dictionaries were
retrained from scratch under two additional seeds (124, 125; identical
data, budget, and $K$): the per-head CS rankings from the three seeds
agree at pairwise Spearman $\rho \geq 0.989$
($0.997$/$0.990$/$0.989$ for the three seed pairs), so the selected
head set is unchanged. On the LVLM towers, every dictionary was
trained twice from independent runs and the selections recomputed:
Gemma recovers the same head specification exactly on two of four
evaluation splits, with a single-head jitter at one layer on the other
two (the same jitter site in both runs); Qwen is exact on three of
four splits with one head differing on the fourth. InternVL marks the
boundary of this stability: the per-layer Jaccard overlap between its
two runs' selections falls from $1.0$ at the shallowest selected layer
to $0.25$ at the deepest --- and this is also the tower on which the
intervention has no effect (\Cref{tab:lvlm_adaptive}), consistent
with the reading that where the selection is not stable, there is no
localized circuit to find. Raw per-head score vectors and comparison
records for all towers accompany the code-and-data supplement.

\section{Threat Model and Adaptive Attacks}
Our defense mutes a \emph{fixed} set of script-sensitive heads; it is a
static defense against a static attack family. The content-sensitivity comparisons of
\Cref{supp:contentcontrols} show the selected heads respond to rendered
script form generally (scrambled glyphs rank-correlate $0.98$/$0.79$
with real words), so attacks that merely change wording or font family
remain within the detected family; attacks that move text \emph{outside}
the script-form distribution (e.g., extreme warping, handwriting, very
low contrast) or that target the unselected middle of the head
distribution are not covered by our evaluation and are plausible
adaptive strategies. Separately, muting script-sensitive heads by design
trades away legitimate text perception (scene-text reading); quantifying
that capability cost on OCR-style probes is left to future work (the
intervention hook in the code supplement exposes the required
machinery). These are disclosed limitations, not claims.

\section{Adaptive Head Gating}
\label{supp:gating}
Static muting acts unconditionally, including on clean inputs. An
input-adaptive variant keeps the operator fixed and makes only the
per-image \emph{decision} adaptive: for each selected head, precompute
the mean $\mu$ and standard deviation $\sigma$ of its CLS
output-contribution norm
$\eta = \| W^O_{(i)} \textstyle\sum_t a_t \mathbf{v}_t \|_2$ on a
clean held-in corpus (no labels), and at test time mute the head for a
given image only when it fires beyond its clean statistics,
$(\eta - \mu)/\sigma > \theta$, applying the $\alpha{=}1$
reallocation when triggered and leaving the forward pass untouched
otherwise. Clean inputs rarely trigger the gate, so clean cost is near
zero by construction; $\theta$ is tuned on held-in data only.

\Cref{tab:gate_matrix} reports the full matrix — five models, both
selections, tuned $\theta^{*}$ per configuration — under the tuning-split protocol ($\theta$ tuned on the
even-index half of ImageNet-100-Text-D; IN100-T$^{\dagger}$ reports the
held-out half; clean on the evaluation half).
\begin{table}[!htbp]
    \centering
    \small
    \setlength{\tabcolsep}{3.6pt}
    \tablebox{%
    \begin{tabular}{llrrrrrr}
        \toprule
        Model & Variant & $\theta^{*}$ & RTA & Disent. & PAINT & IN100-T$^{\dagger}$ & Clean \\
        \midrule
        \multicolumn{8}{l}{\emph{nTAS selection}} \\ \midrule
        ViT-B/16 & Static & -- & $+12.3$ & $+40.6$ & $+11.8$ & $+20.0$ & $-1.2$ \\
         & Gated & $0.25$ & $+11.1$ & $+40.0$ & $+11.8$ & $+19.6$ & $-0.6$$^{\circ}$ \\
        \midrule
        ViT-L/14 & Static & -- & $+13.1$ & $+18.1$ & $+4.5$ & $+16.1$ & $-0.0$$^{\circ}$ \\
         & Gated & $1.0$ & $+11.0$ & $+16.2$ & $+2.7$ & $+16.0$ & $+0.0$$^{\circ}$ \\
        \midrule
        ViT-H/14 & Static & -- & $+22.2$ & $+40.0$ & $+27.3$ & $+22.6$ & $-0.4$$^{\circ}$ \\
         & Gated & $0.5$ & $+21.3$ & $+38.1$ & $+25.5$ & $+21.5$ & $-0.1$$^{\circ}$ \\
        \midrule
        ViT-g/14 & Static & -- & $+15.7$ & $+27.5$ & $+19.1$ & $+19.6$ & $-0.2$$^{\circ}$ \\
         & Gated & $0.25$ & $+13.5$ & $+20.6$ & $+18.2$ & $+18.7$ & $-0.1$$^{\circ}$ \\
        \midrule
        ViT-bigG/14 & Static & -- & $+14.9$ & $+24.4$ & $+20.0$ & $+18.5$ & $-0.3$$^{\circ}$ \\
         & Gated & $0.25$ & $+14.4$ & $+21.9$ & $+19.1$ & $+17.8$ & $-0.2$$^{\circ}$ \\
        \bottomrule
        \midrule
        \multicolumn{8}{l}{\emph{CS selection}} \\ \midrule
        ViT-B/16 & Static & -- & $+9.7$ & $+40.0$ & $+8.2$ & $+19.2$ & $-0.6$$^{\circ}$ \\
         & Gated & $1.0$ & $+9.9$ & $+38.8$ & $+8.2$ & $+18.9$ & $-0.2$$^{\circ}$ \\
        \midrule
        ViT-L/14 & Static & -- & $+11.6$ & $+23.8$ & $+6.4$ & $+14.5$ & $-0.6$$^{\circ}$ \\
         & Gated & $1.5$ & $+10.1$ & $+21.2$ & $+4.5$ & $+14.5$ & $+0.1$$^{\circ}$ \\
        \midrule
        ViT-H/14 & Static & -- & $+21.8$ & $+46.9$ & $+25.5$ & $+23.4$ & $-0.6$$^{\circ}$ \\
         & Gated & $1.0$ & $+18.4$ & $+38.1$ & $+20.9$ & $+21.3$ & $+0.2$$^{\circ}$ \\
        \midrule
        ViT-g/14 & Static & -- & $+17.0$ & $+32.5$ & $+18.2$ & $+19.9$ & $-0.8$ \\
         & Gated & $0.25$ & $+15.0$ & $+27.5$ & $+17.3$ & $+19.1$ & $+0.1$$^{\circ}$ \\
        \midrule
        ViT-bigG/14 & Static & -- & $+13.8$ & $+40.6$ & $+20.9$ & $+12.0$ & $-2.9$ \\
         & Gated & $1.5$ & $+9.9$ & $+31.9$ & $+15.5$ & $+14.4$ & $-0.4$$^{\circ}$ \\
        \bottomrule
        \end{tabular}
        }
    \caption{\textbf{Adaptive gating details.} Five models reported, on both
    selection signals, with tuned $\theta^{*}$ per configuration, under
    the tuning-split protocol. Cells are accuracy deltas (points)
    against each run's paired un-intervened baseline.}
    \label{tab:gate_matrix}
\end{table}

\noindent Columns: selection signal and variant (Static = always-on
muting; Gated = the input-conditional gate), the tuned threshold
$\theta^{*}$ ($\sigma$ units; ``--'' for Static, which has no
threshold), then the accuracy change in points against the paired
un-intervened baseline on the four attack benchmarks and clean
ImageNet-100. $^{\dagger}$held-out half of the tuning benchmark.
$^{\circ}$clean-cost $95\%$ CI contains zero — true for the gated variant
in all ten configurations (both selections, all five models); the static
variant's clean cost excludes zero on three configurations (B/16-nTAS,
g/14-CS, bigG/14-CS). Tuned thresholds $\theta^{*}$ per configuration are listed in the
$\theta^{*}$ column of \Cref{tab:gate_matrix} and recorded in the
evaluation-run manifests; the main paper's Adaptive rows re-evaluate
the same $(\theta^{*}, \text{statistics})$ under the full-benchmark
protocol of the main table.  On B/16 the gate matches static muting on every
attack benchmark at zero measured clean cost; on H/14 it retains most
of the recovery. A patch-level adaptive variant probing a single mean
lexical direction was also evaluated and does not recover attack
accuracy (its per-head mean direction washes out over word-selective
latents; localization AUC $0.37$--$0.65$, versus $0.74$--$0.92$ for
per-latent probes), which is why the head-level gate is the variant
reported.



\section{CKA Context: Late-Layer Head Decorrelation}
\label{supp:cka}
\begin{figure}[!htbp]
\centering
\includegraphics[width=0.48\linewidth]{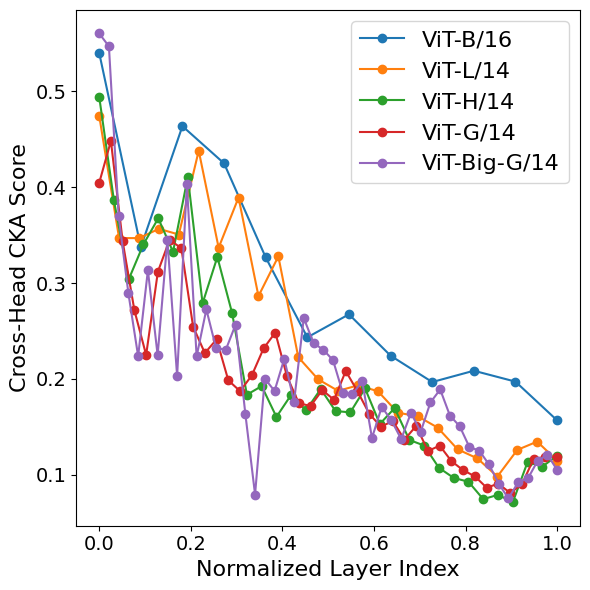}\hfill
\includegraphics[width=0.48\linewidth]{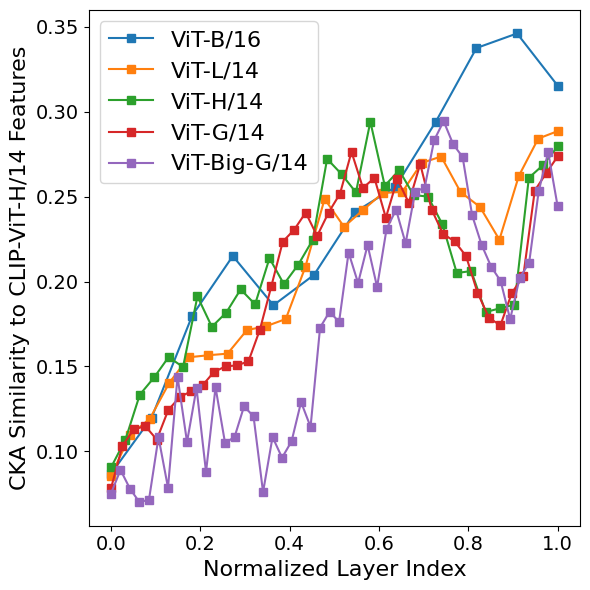}
\caption{\textbf{CKA across normalized depth.} The horizontal axis is normalized depth
(block index divided by model depth, so all five backbones share one
axis). Left: mean pairwise CKA between the output representations of
the heads within a block --- falling values toward the right mean
late-layer heads compute increasingly dissimilar functions. Right:
CKA between each head's output and the model's final \texttt{[cls]}
image embedding --- rising values mean individual late heads
each carry final-representation content. Together: late
layers host specialized, individually influential heads, the
regime the per-head dictionary is built for.}
\label{fig:suppcka}
\end{figure}

Centered Kernel Alignment (CKA; Kornblith et al., 2019) across normalized
ViT depth provides descriptive context for the per-head decomposition:
in the last $20\%$ of CLIP ViT blocks, attention heads grow mutually
decorrelated (\Cref{fig:suppcka}, left), while individual head
outputs align more closely with the final image \texttt{[cls]} embedding
(right). We treat this as context rather than primary motivation: the
residual stream is additive, so later-block alignment with the final
\texttt{[cls]} embedding is partially expected by construction. The
primary evidence that the per-head unit is the right granularity is the
equal-capacity ablation against a monolithic dictionary
(\Cref{supp:arch}).

\section{Computational Cost and Infrastructure}
\subsection{Method Cost Comparison}
\label{app:compute}
\Cref{tab:compute} summarizes the per-image compute footprint of each method. Defense-Prefix~\citep{denseprefix} optimizes a learnable prefix and therefore propagates gradients through the full ViT for $E_{\text{ViT}}$ epochs, dominated by the cost of differentiating through the encoder. Dyslexify~\citep{dyslexify} performs greedy circuit search by evaluating one (layer, head) module per training image, scaling its inference cost as $L\cdot H\cdot N_{\text{train}}$. Our pipeline runs a single ViT forward pass per image to extract intermediate representations, then trains the head-decomposed transcoder offline on the cached activations; once trained, circuit mining only consumes the precomputed nTAS and CS scores and incurs no further ViT compute. The head-wise factorization in \maincref{sec:method_transcoder}{the transcoder section} further reduces the encoder parameter count by $h$ times.

\begin{table}[!htbp]
    \centering
    \caption{\textbf{Computational cost.} Per-image cost decomposed across ViT forward (Fwd), ViT backward (Bwd), and head-decomposed transcoder forward+backward (Dict.\ F+B). $N_{\text{train}}$ is the size of the training set; $E_{\text{ViT}}$ and $E_{\text{dict}}$ are the training epochs of the ViT and the dictionary, respectively; $L$ and $H$ are the number of ViT layers and per-layer attention heads.}
    \renewcommand{\arraystretch}{1.0}
    \resizebox{\linewidth}{!}{
    \begin{tabular}{@{}l c c c@{}}
    \toprule
     & \textbf{ViT Fwd} & \textbf{ViT Bwd} & \textbf{Dict.\ F+B} \\
    \midrule
    Defense-Prefix~\citep{denseprefix} & $E_{\text{ViT}}\cdot N_{\text{train}}$ & $E_{\text{ViT}}\cdot N_{\text{train}}$ & --- \\
    Dyslexify~\citep{dyslexify}        & $L\cdot H\cdot N_{\text{train}}$       & ---                                   & --- \\
    Ours                               & $N_{\text{train}}$                     & ---                                   & $E_{\text{dict}}\cdot N_{\text{train}}$ \\
    \bottomrule
    \end{tabular}
    }
    \label{tab:compute}
\end{table}

\subsection{Infrastructure and Evidence Map}
All experiments ran on a single node with 8$\times$ NVIDIA RTX A6000
(48\,GB) GPUs, 503\,GB RAM, Linux 5.15; PyTorch 2.7.0+cu126,
open\_clip, pytorch-lightning, numpy/scipy (exact versions pinned in the
code supplement's environment file). Dictionary training uses cached
activations; all evaluation is inference-only. 


\subsection{Signal Degradation from Concept Entanglement}
\label{app:theory}

\paragraph{Encoder Definitions.}
Let a latent~$c$ have a \emph{head support} $\mathcal{H}_c \subseteq \{1,\ldots,h\}$, defined as the set of attention heads whose value subspaces contribute nonzero weight to $c$'s encoder direction. We contrast two architectures based on this support:
\begin{itemize}
    \item \textbf{Monolithic encoders} map the concatenated output of all $h$ attention heads to latent activations. Nothing in this architecture constrains the size of the support, allowing $|\mathcal{H}_c| \geq 1$.
    \item \textbf{Per-head encoders} isolate feature learning by restricting each latent to a single head's value subspace, guaranteeing $|\mathcal{H}_c| = 1$ by construction for every latent.
\end{itemize}

\paragraph{Assumption: Head Division of Labor.}
We assume the underlying transformer distributes distinct functional mechanisms across different attention heads. Formally, let a generalized semantic concept~$c$ comprise orthogonal sub-features (e.g., visual versus lexical mechanisms). Under the head division of labor, these structurally distinct sub-features are isolated within non-overlapping heads $i$ and $k$. Consequently, to successfully represent the full semantic breadth of~$c$, the monolithic encoder's optimal projection must draw from these disjoint specialized heads, structurally forcing its support size to be strictly greater than one: $|\mathcal{H}_c| \geq 2$.

\paragraph{Activating Set Contamination and Clarity Score.}
The interpretability of a latent is fundamentally tied to its \emph{activating set}: the subset of tokens across a dataset that cause the latent to fire. In a monolithic architecture, because latent~$c$ is a mixture of orthogonal mechanisms, its activating set becomes contaminated. The latent will fire both when head~$i$ detects a visual shape and when head~$k$ detects a lexical string. 

Consequently, the activating set for latent~$c$ is a heterogeneous mixture of functionally distinct tokens. Because the Clarity Score (CS) evaluates the purity and quality of the conceptual cluster formed by a latent's activations, this contamination severely degrades the score. The latent fails to carve out a clean, monosemantic cluster, rendering it mathematically polysemantic under CS evaluation.

\paragraph{Per-Head Resolution.}
The per-head encoder resolves this contamination by enforcing $|\mathcal{H}_c| = 1$. By strictly preventing cross-head signal summation during training, distinct semantic concepts cannot be optimized into the same latent space, even under sparsity pressure. A per-head latent tracks either the visual mechanism of head~$i$ \emph{or} the lexical mechanism of head~$k$, but never both. Consequently, its activating set remains strictly homogeneous. This structural isolation guarantees a pure conceptual cluster, preserving the latent's Clarity Score and ensuring true monosemanticity.

\end{document}